\pdfoutput=1

\documentclass[11pt]{article}

\usepackage[]{ACL}

\usepackage{times}
\usepackage{latexsym}
\usepackage{enumitem}
\usepackage{float}

\usepackage[T1]{fontenc}

\usepackage[utf8]{inputenc}

\usepackage{microtype}
\usepackage{subcaption}

\usepackage{inconsolata}
\usepackage[size=tiny]{todonotes}
\newcommand{\se}[1]{\textcolor{black}{#1}}
\newcommand{\senew}[1]{\textcolor{black}{#1}}
\newcommand{\senewest}[1]{\textcolor{black}{#1}}
\newcommand{\cl}[1]{\textcolor{black}{#1}}
\newcommand{\revision}[1]{\textcolor{black}{#1}}
\newcommand{\green}[1]{\textcolor{green}{#1}}
\newcommand{\gray}[1]{\textcolor{gray}{#1}}
\newcommand{\cll}[1]{\textcolor{black}{#1}}
\newcommand{\rev}[1]{\textcolor{black}{#1}}
\newcommand{\revf}[1]{\textcolor{black}{#1}}
\newcommand{\revn}[1]{\textcolor{black}{#1}}

\newcommand{\blue}[1]{\textcolor{blue}{#1}}
\newcommand{\orange}[1]{\textcolor{orange}{#1}}
\usepackage{newunicodechar}
\usepackage{amssymb}
\usepackage{amsmath}
\usepackage{booktabs}
\usepackage{array}

\newcommand\Warning{%
 \makebox[1.4em][c]{%
 \makebox[0pt][c]{\raisebox{.1em}{\small!}}%
 \makebox[0pt][c]{\color{red}\Large$\bigtriangleup$}}}%

\newunicodechar{⚠}{\Warning}

\title{PrExMe! 
Large Scale Prompt Exploration of Open Source LLMs for \\ Machine Translation and Summarization Evaluation 
}

\author{
  \textbf{Christoph Leiter\textsuperscript{1}},
  \textbf{Steffen Eger\textsuperscript{1,2}}
  \\
Natural Language Learning Group (NLLG)\\ \url{https://nl2g.github.io/}\\
\\
  \textsuperscript{1}University of Mannheim,
  \textsuperscript{2}University of Technology Nuremberg  (UTN)\\
  \small{
    \textbf{Correspondence:} \href{mailto:email@domain}{christoph.leiter@uni-mannheim.de}, \href{mailto:email@domain}{steffen.eger@utn.de}
  }
}

\begin{document}
\maketitle
\begin{abstract}
Large language models (\textsc{LLMs}) have revolutionized \textsc{NLP} \revn{research}. \revn{Notably, in-context learning enables} their use as evaluation metrics for natural language generation, 
making them particularly advantageous in low-resource scenarios and time-restricted applications. 
In this work, we introduce PrExMe, a large-scale \textit{\textbf{Pr}ompt \textbf{Ex}ploration for \textbf{Me}trics}, where we evaluate more than $720$ prompt templates for open-source \textsc{LLM}-based metrics on machine translation (\textsc{MT}) and summarization datasets, totalling over $6.6$M evaluations. This extensive comparison (1) 
\revn{benchmarks} recent open-source \textsc{LLMs} as metrics and (2) explores the stability and variability of different prompting strategies. We discover that, on the one hand, there are scenarios for which prompts are stable. For instance, some \textsc{LLMs} show idiosyncratic preferences and favor to grade generated texts with textual labels while others prefer to return numeric scores. 
On the other hand, the stability of prompts and model rankings can be susceptible to seemingly innocuous changes. 
For example, changing the requested output format from ``0 to 100'' to ``-1 to +1'' can strongly affect the rankings in our evaluation. 
Our study contributes to understanding the impact of different prompting approaches on \textsc{LLM}-based metrics for MT and summarization evaluation, highlighting the most stable prompting patterns and potential limitations.\footnote{We make our code available: \url{https://github.com/Gringham/PrExMe}}
\end{abstract}

\section{Introduction} \label{sec:introduction}
The recent  
success of 
\senew{\textsc{LLMs}}
\revn{has} led to a paradigm shift in  
\senew{\textsc{NLP}} 
\citep{zhang2023nllg}. Instruction-tuning allows \textsc{LLMs} to \revn{respond}  
to complex task descriptions (prompts) \citep{NEURIPS2022_b1efde53}, \revn{including} 
conventional \textsc{NLP} tasks, \revn{like automatically evaluating} \cl{natural language generation (\textsc{NLG})}  
\revn{for } machine translation (\textsc{MT}) and summarization.
\begin{figure}[!ht]
\includegraphics[width=0.48\textwidth]{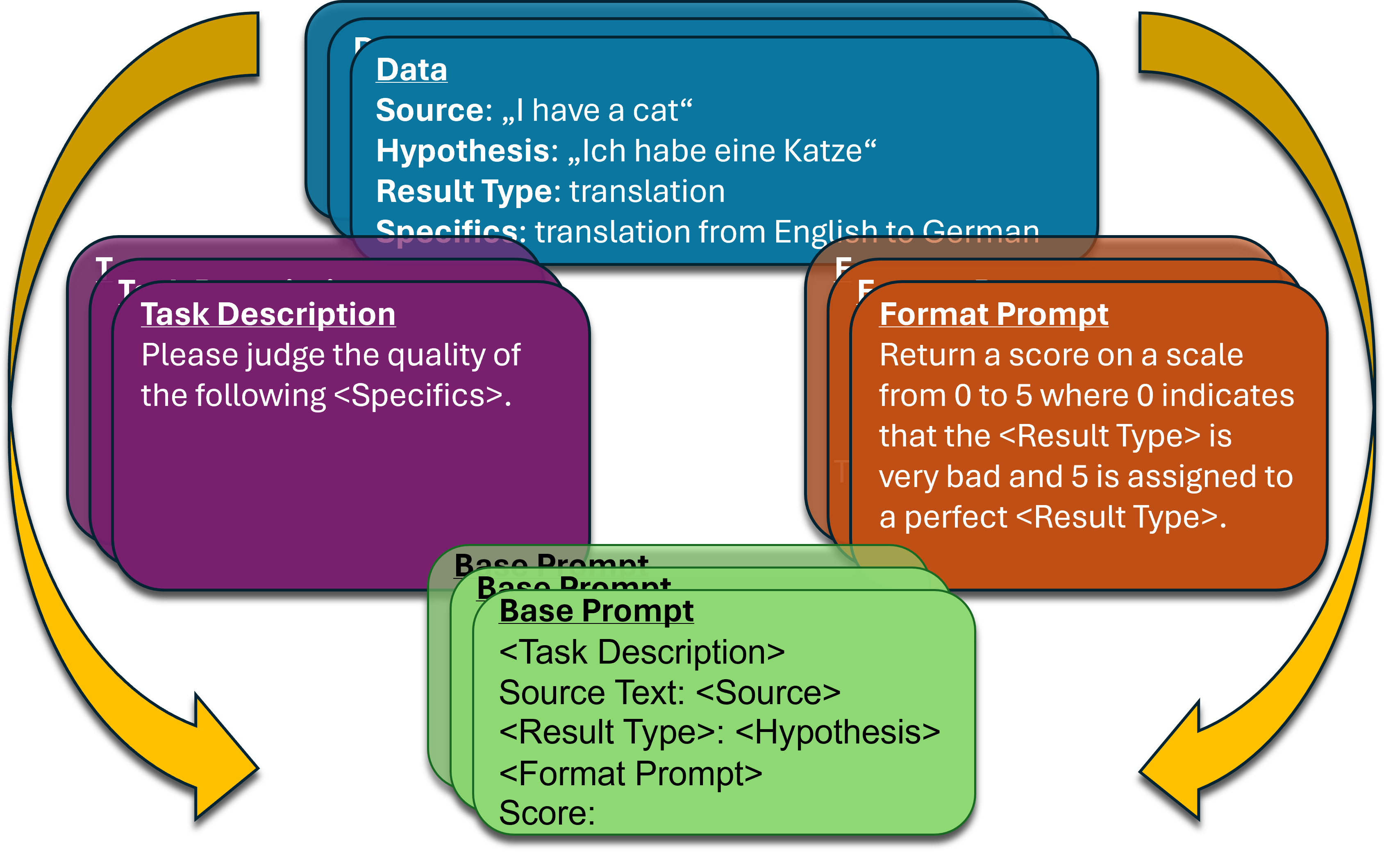}
\caption{Schematic overview of our prompt exploration methodology, featuring a grid search across \textit{datasets}, \textit{task descriptions}, \textit{output formats}, and \textit{base prompts}.}
\label{PrEx}
\end{figure}

\revn{Building on this}, researchers \revn{increasingly} use \revision{\textsc{LLMs} as evaluation metrics, \revn{achieving} remarkable performance, sometimes relying solely} on in-context learning \citep[e.g.][]{kocmi-federmann-2023-gemba,fernandes-etal-2023-devil}, i.e., \revn{with}
metrics that \revn{only use} prompting.  
Such prompting-based metrics require 
\revn{minimal data}, 
\revision{making them useful for low-resource evaluation scenarios 
\citep{belouadi-eger-2023-uscore}  
\revn{and} more resource-efficient since they do not require fine-tuning.}

Although many prompting-based metrics have been proposed \citep[e.g.][]{li2024leveraging}, structured evaluations across different prompting approaches remain scarce, especially for open-source models. \revn{The recent} \textsc{Eval4NLP 2023} shared task \citep{leiter2023eval4nlp} addresses this  
by (1) restricting the usage to selected open-source LLMs and (2) prohibiting their fine-tuning. 
While the \senew{shared-task} submissions \revn{offer interesting insights}, they focus on only a few distinct prompts\revn{, leaving the impact and robustness of prompt variations  
largely unexplored.}

In this work, we introduce a  
\se{systematic} \textbf{Pr}ompt \textbf{Ex}ploration for \textbf{Me}trics \se{(PrExMe)}, 
\revn{expanding on} \textsc{Eval4NLP 2023} to provide a much larger, template-based, structured evaluation of the effects different input prompts have on an LLM-based metric's correlation \revision{with} human judgements in MT and summarization evaluation. We formulate the following research questions:
\begin{enumerate}
    \itemsep-0.1em 
    \item[RQ1] Can open-source \textsc{LLMs} evaluate text generation without fine-tuning \revision{and how do they differ from each other} (see \S\ref{rq1})?
    \item[RQ2] Can we identify patterns\footnote{\revision{We define \textit{prompting patterns} as the template components that constitute a prompt (e.g., zero-shot, one-shot or the output format).}} in prompts that lead to a stable performance across different datasets, tasks, and models (see \S\ref{rq2})?
    \item[RQ3] How should researchers design prompts for new evaluation scenarios (see \S\ref{rq3})?
\end{enumerate}

\revn{In PrExMe, we construct} hierarchical templates based on approaches \revision{such as \textit{chain-of-thought} (\textsc{CoT}) \citep{NEURIPS2022_8bb0d291}, \textit{zero-shot} and \textit{retrieval-augmented generation} (\textsc{RAG})} \citep{gao2024retrievalaugmented}. Each template gets filled with further sub-templates. For example, we vary the requested output formats, such as numeric \revision{\textit{scores} and \textit{textual labels} (see \S\ref{sec:setup})}. 
This setup amounts to \revision{more than} 720 
templates that we evaluate with 7 \textsc{LLMs} \revn{in a 1st phase}. In a  
\senew{2nd}
phase, we test the generalizability and performance of the prompts with the best correlations on two further datasets. \revn{W}e make the following key contributions:  
\begin{enumerate} [noitemsep, left=0pt] 
\itemsep-0.1em

    \item[\green{\checkmark}] We \revn{conduct} a large\se{-}scale analysis \revn{of} over 6.6M\ prompts  
    \revn{for} \textsc{LLM}-based metrics \revn{in} \textsc{MT} and summarization evaluation. \revision{This \revn{thorough} exploration \revn{covers} various prompting \revn{techniques}, datasets, tasks, and models, making it, to our knowledge, the most extensive  
    of its kind.} 
    \item[\green{\checkmark}] \revision{We show that certain prompting patterns are robust and generalizable across different \rev{tasks} and datasets, with median performance being a good predictor for new settings. \revn{For example, some models show a distinctive preference for textual labels, while others yield better results with numeric labels.} However, in some cases, even minor prompt changes can significantly impact performance.}
    \item[\green{\checkmark}] \revision{Our study tackles prompt-based evaluation with open-source LLMs, targeting scenarios where fine-tuning or access to closed-source LLMs is not possible. Such evaluations are still very scarce but important to make research more accessible, \senew{fostering diversity and inclusion}.} 
    \item[\green{\checkmark}] \revn{We} systematically test established prompting approaches, including zero-shot, CoT and RAG, to comprehensively evaluate the performance of recent open-source LLMs for evaluation metrics. \revf{Aligning with the recommendations of \citet{mizrahi2024state}, we use multiple prompts per model which mitigates the risk of single prompts disproportionately affecting their performance and ensures a fair comparison.} The \textsc{Platypus2-70B} model \citep{lee2023platypus} achieves the strongest performance among the tested LLMs.
\end{enumerate}

\section{Related Work} \label{sec:related}
\revn{We first discuss related work on prompting-based metrics for MT and summarization, then connect our study to research on prompting techniques and stability.}

\paragraph{Prompting-based metrics}
Recent advances in \textsc{LLM}-based metrics for \textsc{NLG} often rely on in-context learning \revn{to predict the quality of generated texts}. Surveys by \citet{li2024leveraging} and \citet{gao2024llmbased} \revn{offer} comprehensive overviews of these metrics.
\revn{Such prompt-based approaches are often built upon closed-source models or test only a few prompts. For example, GEMBA \citep{kocmi-federmann-2023-large}, GEMBA-MQM \citep{kocmi-federmann-2023-gemba} and AutoMQM \citep{fernandes-etal-2023-devil} evaluate strong prompting approaches for MT evaluation with closed-source models. \citet{lu2024erroranalysispromptingenables} use ChatGPT  \citep{chatgpt} and two open-source models, to explore one novel prompting approach.}  In contrast, the \textsc{Eval4NLP} 2023 shared task \citep{leiter2023eval4nlp}, considers 
open-source prompt-based metrics, where participants evaluate \rev{MT} and \rev{summarization} using only allowed models without fine-tuning.  
While Eval4NLP yielded interesting techniques, the participants explored a limited range of prompts, leaving a gap in the comprehensive analysis of prompting patterns and \revn{consistent \textsc{LLM} comparisons.} 

\revn{PrExMe addresses these research gaps by systematically analyzing a much larger set of prompts across comparable experiment\se{al} settings to (1) study the robustness of prompts across datasets, open-source} models and tasks, and to (2) \revn{identify patterns to guide future prompt-based metrics.}

\paragraph{Prompting Techniques} 
\revision{
\revn{Over recent years, many successful prompting techniques have been developed} \citep[e.g.,][]{10.1145/3560815}. Our work primarily builds on established methods like Zero-Shot CoT and \senew{RAG}. 
Further, \citet{Li2023LargeLM} propose emotion-inducing prompts to improve LLM performance. 
\senew{T}o our best knowledge, we are the first to analyze \rev{this technique} for evaluation metrics. Inspired by this, we also propose a novel emotion-CoT pattern (see \S\ref{sec:setup}). \citet{kocmi-federmann-2023-large} previously evaluated output formats for prompt-based metrics, which we extend with a much broader analysis. } 
\rev{O}ther works use hierarchical templates for prompt building \citep[e.g.][]{fu2023gptscore} and tools like LangChain \citep{Langchain} and DSPy \citep{DSPy} support the\rev{ir} 
implementation. We employ hierarchical templates for structured comparisons among prompting patterns.

\paragraph{Prompting Robustness} 
\revn{Our grid search across various prompts, datasets and tasks extends research on} how \textsc{LLMs} \revision{respond to prompt perturbations}.  
\citet{webson-pavlick-2022-prompt}, \citet{leidinger-etal-2023-language}, \citet{weber2023icl} and \citet{sclar2023quantifying} reveal significant performance variations in tasks like natural language inference and sentiment classification. As a solution, \citet{sclar2023quantifying} suggest \revn{reporting the full range of results across prompt perturbations, while \citet{voronov2024mind} and \citet{mizrahi2024state} argue for using multiple templates to increase the reliability of evaluation benchmarks. }
To our \cl{be}\se{s}\cl{t} knowledge, \senew{we are} 
the first to explore to which degree these robustness \revn{issues} affect open-source \textsc{LLM}-based metrics \rev{and how \senewest{to} select the best prompts. Also, by prompting the LLMs with multiple prompts, we follow \citet{mizrahi2024state} and achieve a stable and fair evaluation of LLMs as MT and summarization metrics.}

\section{Setup} \label{sec:setup}
\rev{In this section, we present the templates and prompting techniques 
we use to utilize \textsc{LLMs} as metrics, and we \revn{outline} the datasets and models that we use for testing.} We evaluate \textsc{LLMs} in a \se{\textbf{reference-free}} setting \revn{(}grading a generated hypothesis 
based on its source without a 
reference\revn{)}.\footnote{We run experiments using \textsc{vLLM} \citep{kwon2023efficient} \revn{with greedy decoding} on two clusters with Nvidia A6000, A40 and A100 GPUS.  
\senew{D}etails on versions, tools and model parameters 
\senew{are} 
in Appendix \ref{sec:metric_details}.}
\cll{The evaluated prompt \revn{templates} provide a comprehensive evaluation framework for LLM-based metrics, \revn{covering} basic in-context learning, sophisticated reasoning, emotional context, and varying output structures, ensuring a thorough 
assessment of robustness and adaptability across tasks and datasets.}

\paragraph{Prompt Templates} \revn{We construct prompts as hierarchical templates (see Figure \ref{PrEx}), with large templates built from 
smaller ones}. 
\revision{Each prompt is \revn{built} from: (1) the \textit{source text} and generated \textit{hypothesis text} that should be graded\senewest{,} (2) a \textit{base prompt}, (3) a \textit{task description}, (4) a \textit{format requirement} and (5) optionally a one-shot \textit{demonstration}. Table \ref{tab:CombinedTable} presents examples for (2), (3), (4) and (5).}

\begin{table*}
\small
    \centering
    \begin{tabular}{|l|p{0.83\linewidth}|}
        \hline
        \textbf{Category} & \textbf{Description} \\
        \hline
        \textbf{Base Prompts} & 
        \textbf{\textsc{PZS}}: ``\{task\_description\} \textbackslash nSource Text: \{src\} \textbackslash n\{result\_type\}: \{hyp\} \textbackslash n\{format\_requirement\} \textbackslash nScore: '' \\
        & \textbf{\textsc{ZS-CoT-EM}}: ``\{task\_description\} \textbackslash nSource Text: \{src\} \textbackslash n\{result\_type\}: \{hyp\} \textbackslash n\{format\_requirement\} \textbackslash nFirst describe your emotions, then think step by step and explain your thought process, finally return your judgment in the format 'Judgment: '.'' \\
        & \textbf{\textsc{OS-CoT}}: ``\{task\_description\} \textbackslash n Here is an example:\textbackslash n Source Text: \{ex\_src\} \textbackslash n\{result\_type\}: \{ex\_hyp\}\textbackslash n Judgement: <Description of reasons>. Therefore the score is \{ex1\_score\}\textbackslash n\textbackslash n Now it is your turn to grade the \{result\_type\}.\textbackslash n Source Text: \{src\} \textbackslash n\{result\_type\}: \{hyp\} \textbackslash n\{format\_requirement\} \textbackslash n First, think step by step and explain your thought process, then return your judgment in the format 'Judgment: '.'' \\
        \hline
        \textbf{\rev{Task Desc.}} & 
        \textbf{Neutral}: ``Judge the quality of the following \{task\_specific\_insert\}.'' \\
        & \textbf{Sceptical}: ``I'm not sure about this one. Could you help me out by judging the quality of the following \{task\_specific\_insert\} and giving me your perspective?'' \\
        & \textbf{Empathetic}: ``I know it isn’t an easy task, but it would be really great of you if you could help me judge the quality of the following \{task\_specific\_insert\}.'' \\
        \hline
        \textbf{Format Req.} & 
        \textbf{0 or 1}: Return a discrete score of 0 if the \{result\_type\} has flaws and 1 if it is perfect. \\
        & \textbf{catastrophic}, \textbf{indifferent} or \textbf{marvelous}: Choose whether the \{result\_type\} is either "catastrophic", "indifferent" or "marvelous". \\
        \hline
    \end{tabular}
    \caption{\revn{P}rompt templates for the \textit{base prompt}, \textit{task description}, and \textit{format requirements} \revn{(Full list: Appendix \ref{sec:prompt_templates})}.}
    \label{tab:CombinedTable}
\end{table*}

\revision{The \textbf{base prompt} is the top layer of our prompt hierarchy, incorporating the other components. \revn{We} test three zero-shot} (ZS) and corresponding one-shot (OS) base prompts: (1) \textit{Plain ZS/OS} (\textsc{PZS/POS}), (2) \textsc{ZS/OS-CoT} and (3) \textit{ZS/OS-CoT-Emotion} (\textsc{ZS/OS-CoT-EM}). 
\textsc{PZS} plainly presents the newline-separated \textit{task description}, \textit{source}, \textit{hypothesis} and \textit{format requirement}. \textsc{ZS-CoT \citep{NEURIPS2022_8bb0d291}} asks the model to \textit{think step by step} before returning its output. Lastly, \textsc{ZS-CoT-EM} 
asks the model to describe its ``emotions'' before the ZS-CoT prompt. 
We include \textsc{CoT} due to its success in enhancing prompt-based performance in metrics like \textsc{AutoMQM} \citet{fernandes-etal-2023-devil}, EAPrompt \citet{lu2024erroranalysispromptingenables} and \textsc{GEMBA} \citep{kocmi-federmann-2023-gemba}. \textsc{ZS-CoT-EM} examines LLM performance variations when prompted to express emotions, inspired by our exploration of emotional prompts (see ``task description'' below). The OS versions of the templates add a \revn{demonstration field}. To avoid fixating the model on specific reasoning steps, we include a placeholder for OS-CoT where the model should insert its reasoning.

The \textbf{task description} \revn{is the} instruction to grade the generated hypothesis. \citet{Li2023LargeLM} 
find that instructions \revn{evoking} certain emotions for humans \revn{can enhance \textsc{LLM} performance}. Inspired by this, we \revn{experiment with} ``emotional prompts'' in the task description. \revn{This approach primarily broadens our grid search through simple paraphrasing but also allows us to study the effect of emotions on \textsc{LLM}-based metrics.} Besides \textit{neutral} prompts, we include instructions like \textit{polite}, \textit{threatening} and \textit{sceptical}. We create 11 task descriptions ourselves and 13 further descriptions with \textsc{ChatGPT}.

\revn{The \textbf{format requirement} specifies the output format the \textsc{LLM} should follow when generating a score, including the score range and whether it should be discrete or continuous. We also include prompts that ask the \textsc{LLM} to return textual quality labels. Overall, we define 10 format requirements.}

Lastly, we construct the optional OS \textbf{demonstrations} with 
\senew{RAG}\senew{.}  
We extract demonstrations from \textsc{WMT21} \citep{freitag-etal-2021-results} for MT and \revn{the factuality dataset} \textsc{RoSE} for summarization \citep{liu-etal-2023-revisiting}. For each demonstration sample in both datasets \revision{and for each input sample of our metric}, we create sentence embeddings with \textsc{XLMR-SBERT} \citep{reimers-gurevych-2020-making}. \revn{Thereby, we concatenate the source and hypothesis embeddings for the input samples}. We then select the demonstration with the highest cosine similarity for each input. \revision{Due to resource constraints, we evaluate only the 9 best ZS prompts in an OS setting, as described in the \textit{Datasets and phases} section below.}

\paragraph{MQM-based approaches}
Additionally to hierarchical templates, we test the \textsc{GEMBA-MQM} prompts \citep{kocmi-federmann-2023-gemba} with our selected open-source \textsc{LLMs}. \textsc{GEMBA-MQM}, which predicts scores based on the number of present errors weighted by severity, normally uses \textsc{GPT4}. We refer to \rev{the open-source implementation} as \textit{LocalGemba}.

\paragraph{Score Extraction \& Evaluation}
We restrict generation to 180 tokens and extract the last regex match of a label or \revn{any} number as the score.  When no result is found\senewest{,} we average the other scores of its prompt template. During evaluation, we map textual labels to 1, 3 and 5. 

We evaluate prompt templates at the segment-level, like the WMT QE and metrics shared tasks \citep[e.g.][]{freitag-etal-2022-results,freitag-etal-2021-results,zerva-etal-2022-findings}. \revision{That means, for each metric we compute the correlation between metric scores and ground truth human judgments without averaging by system or document.} 
As correlation measures, we use the Kendall \revn{(primary measure)} \citep{10.1093/biomet/33.3.239}, Pearson and Spearman correlations, as well as tie-calibrated accuracy \citep{deutsch-etal-2023-ties}. Further, we compute permute-input significance tests ($p\le 0.075$) \citep{deutsch-etal-2021-statistical} for the Kendall correlations presented in our result \rev{t}ables. \revn{Since often no single performance is significant, we report clusters where each included metric significantly outperforms those excluded.}

\begin{table*}[htb]
\small
    \centering
     \begin{tabular}{|l|l|l|l|l|l|l|l|l|l|l|l|}
        \hline
        & \multicolumn{3}{|c|}{\textbf{P1: Eval4NLP train}} & \multicolumn{4}{|c|}{\textbf{P2: Eval4NLP test}} & \multicolumn{4}{|c|}{\textbf{P2: WMT23/Seahorse}} \\
        \hline
        \textbf{Model} & \textbf{en-de} & \textbf{zh-en} & \textbf{summ} & \textbf{en-de} & \textbf{en-es} & \textbf{en\_zh} & \textbf{summ} & \textbf{en-de} & \textbf{he-en} & \textbf{zh-en} & \textbf{summ} \\
        \hline
        \multicolumn{12}{|c|}{\textbf{\rev{1.\ Hierarchical Templates}}} \\
        \hline
        LL3-70B & \textcolor{blue}{0.273} & \textcolor{orange}{0.306} & \textcolor{orange}{0.442} & \textcolor{orange}{0.245} & \textcolor{orange}{0.189} & \textcolor{orange}{0.231} & \textcolor{orange}{0.438} & \textcolor{orange}{0.297} & \textcolor{orange}{0.172} & \textcolor{orange}{0.312} & \textcolor{orange}{0.312} \\
        LL3-8B & \textcolor{blue}{0.251} & \textcolor{orange}{0.236} & \textcolor{orange}{0.334} & \textcolor{blue}{0.167} & \textcolor{blue}{0.158} & \textcolor{blue}{0.145} & \textcolor{orange}{0.412} & \textcolor{orange}{0.166} & \textcolor{blue}{0.118} & \textcolor{orange}{0.164} & \textcolor{blue}{0.200} \\
        MI-7Bx8 & \textcolor{orange}{0.268*} & \textcolor{orange}{0.264} & \textcolor{orange}{0.365} & - & - & - & - & - & - & - & - \\
        NO-13B & \textcolor{blue}{0.230} & \textcolor{orange}{0.201} & \textcolor{blue}{0.225} & \textcolor{orange}{0.205} & \textcolor{orange}{0.141} & \textcolor{orange}{0.084} & \textcolor{orange}{0.255} & \textcolor{orange}{0.202} & \textcolor{orange}{0.105} & \textcolor{orange}{0.175} & \textcolor{orange}{0.123} \\
        OR-13B & \textcolor{blue}{0.289} & \textcolor{blue}{0.303} & \textcolor{blue}{0.468*} & \textcolor{orange}{0.214} & \textcolor{blue}{0.158} & \textcolor{blue}{0.206} & \textcolor{blue}{0.518} & \textcolor{blue}{0.375} & \textcolor{blue}{0.247} & \textcolor{blue}{0.387} & \textcolor{blue}{0.377} \\
        PL-70B & \textcolor{blue}{\textbf{0.344*}} & \textcolor{blue}{\textbf{0.364*}} & \textcolor{blue}{\textbf{0.519*}} & \textcolor{blue}{\textbf{0.402*}} & \textcolor{blue}{\textbf{0.289*}} & \textcolor{blue}{0.295*} & \textcolor{blue}{0.549} & \textcolor{blue}{0.338} & \textcolor{blue}{\textbf{0.259*}} & \textcolor{blue}{\textbf{0.417*}} & \textcolor{blue}{\textbf{0.448*}} \\
        TO-13B & \textcolor{orange}{0.284*} & \textcolor{orange}{0.318*} & \textcolor{orange}{0.375} & \textcolor{orange}{0.379*} & \textcolor{orange}{0.253} & \textcolor{orange}{0.232} & \textcolor{orange}{0.409} & \textcolor{orange}{0.322} & \textcolor{orange}{0.208} & \textcolor{orange}{0.314} & \textcolor{orange}{0.257} \\
        \hline
        \multicolumn{12}{|c|}{\textbf{\rev{2.\ Separate Prompting Techniques}}} \\
        \hline
        M:LG & 0.278* & 0.268 & 0.062 & 0.344 & 0.265 & \textbf{0.307*} & 0.116 & \textbf{0.391*} & 0.190 & 0.300 & 0.144 \\
        B:DSBA & 0.164 & 0.306 & 0.458 & 0.314 & 0.226 & 0.159 & \textbf{0.600*} & 0.172 & 0.207 & 0.376 & 0.373 \\
        \hline
        \multicolumn{12}{|c|}{\textbf{\rev{3.\ Baselines with External Base Models}}} \\
        \hline
        B:BS & 0.056 & -0.109 & 0.155 & 0.125 & 0.139 & -0.009 & 0.421 & -0.018 & 0.001 & -0.167 & 0.069 \\
        B:XC & \textcolor{gray}{0.629} & \textcolor{gray}{0.513} & -0.069 & 0.468 & 0.298 & 0.387 & 0.224 & 0.531 & 0.300 & 0.447 & 0.146 \\
        \hline
    \end{tabular}
    \caption{\rev{Kendall correlations of the best} performing prompts of the phase 1 (P1) and phase 2 (P2) evaluations across various datasets. \rev{Abbreviations are defined in Appendix \ref{sec:model_abbreviations}.
    Vertically, we group the table into (1) correlations achieved with our \textit{hierarchical templates}, (2) correlations of prompting techniques that are explored \textit{separately} from the hierarchical templates, but use the same base model(s) and (3) baselines that use \textit{external base models}, i.e., that are not based on the same LLMs.
    For each column the \textbf{bold} value indicates the highest correlation and correlations with an asterisk (*) are significantly higher $(p\le0.075)$ than those without (excluding group (3)). The grey values for XC indicate tasks that were included in its training data. }
    The MQM based approach is marked with \textit{M:} and baselines are marked with \textit{B:}. \orange{Orange} values indicate that the prompt required textual quality labels, while \blue{blue} values indicate numeric labels. More details can be found in Appendix \ref{phase2perf}.}
    \label{tab:combined}
\end{table*}

\paragraph{\revision{Models}}
We select instruction-tuned \textsc{LLMs} \revision{with strong} performance in \textsc{Eval4NLP 2023}: \se{(1)} \textsc{Platypus2-70B-Instruct-GPTQ}, 
\se{(2)} \textsc{Nous-Hermes-13b} \footnote{\url{https://huggingface.co/NousResearch/Nous-Hermes-13b}} and 
\se{(3)} \textsc{OpenOrca-Platypus2-13B} 
\citep{hunterlee2023orcaplaty1, mukherjee2023orca}. 
We abbreviate these as \textsc{Platypus2}, \textsc{Nous} and \textsc{Orca}. 
\revision{Additionally, we evaluate 
more recent models: \senewest{(4)} \textsc{LLaMA3-8B} \citep{llama3modelcard}, \senewest{(5)} a GPTQ version of \textsc{LLaMA3-70B} \citep{llama3modelcard}, \senewest{(6)} \textsc{Mixtral-8x7B} \citep{jiang2024mixtral} \revn{(excluded in phase 2 due to resource use and weaker performance)} and \textsc{Unbabel-TowerInstruct} \citep{tower_llm_2024}\senewest{,} a 13B parameter multilingual instruction-tuned model.}

\paragraph{\revision{Datasets and phases}}
We conduct our experiments in two phases on different datasets. By doing so, we want to mitigate statistical effects of our extensive prompt search. Also, it allows  
to evaluate selected prompts on full datasets, 
a task that would otherwise be too resource intensive, 
and to explore generalizability. 

In phase 1, we evaluate on the train set of \textsc{Eval4NLP 2023} \citep{leiter2023eval4nlp}, and in phase 2, on its dev and test sets.  
The train and dev sets are (reference-free) splits of the \textsc{WMT2022} metrics shared task \citep{freitag-etal-2022-results} \revn{for MT} and \textsc{SummEval} \citep{fabbri-etal-2021-summeval} \revn{for summarization}. The test set was newly annotated by \citet{leiter2023eval4nlp} \revn{and also contains human MT and summarization quality annotations}. In phase 2, we also evaluate the ZS prompts\footnote{\revn{As OS prompts performed weakly on the other datasets, we do not evaluate them on WMT23 and Seahorse.}} on the WMT23 MQM annotations for MT \citep{freitag-etal-2023-results} and \textit{Seahorse} \citep{clark-etal-2023-seahorse} for \rev{multilingual} summarization. \revn{The summarization datasets that we evaluate target \textit{overall summary quality}, i.e., human annotations for separate aspects like \textit{coherence} or \textit{fluency} are aggregated into single scores.} More details of the datasets are discussed in Appendix \ref{dataset_details}.

In the \textbf{\senew{1st} phase}, we evaluate all 720\footnote{Considering the different tasks, this number could also be considered higher.} \revision{combinations of \se{ZS} prompts on the Eval4NLP train set}.  As this is 
resource intensive, 
for MT we restrict ourselves to the first 500 samples \revision{of each language pair}. \revision{We then select the prompt with the highest Kendall 
correlation for each \textit{task+base prompt} combination, yielding 9 unique prompts \revn{for phase 2 (see Appendix \ref{app:best_prompts})}. \revn{That means, we select the highest correlating PZS, \textit{ZS-COT} and \textit{ZS-COT-EM} prompt for each of the phase 1 tasks en-de, zh-en and summarization. In case of duplicates, we choose the second highest correlation. While this approach might result in the prompts of stronger models being favored for phase 2, the distribution across different base prompts and tasks is broad enough to enable a comprehensive analysis of each prompting pattern.}}

In the \textbf{\senew{2nd} phase}, we evaluate the selected prompts of the 
\senew{1st} 
phase on all samples of the respective datasets (Eval4NLP dev+test, WMT23 and Seahorse).  
\senew{This} further test\senew{s} the generalizability of prompts between models and for unseen, in-domain data (the Eval4NLP dev set stems from the same original datasets) and out-domain data. 

\paragraph{Baselines}
For each phase, we also present correlations for \cl{two baseline metrics that use other base models: \textsc{BARTScore} \citep{bartscore} and \textsc{XComet} \citep{guerreiro2023xcomet}. Especially \textsc{XComet} has the benefit of being trained on multilingual datasets.} 
Further, we test the prompts of \textsc{DSBA} \citep{kim-etal-2023-better} \cll{--- that showed a strong performance for summarization in the shared task ---} with \rev{Platypus2-70B and Orca-13B}.

\section{Results} \label{sec:results} 
In \textit{phase 1}, we run 6,652,800 ZS prompts \revn{(720 prompt templates with 7 models on 1320 samples)} and 71,280 OS prompts \revn{(9 ``best'' prompt templates)}, with no scores extracted in 12.7\% resp.\  19.4\% of cases; the average score of the prompt template is assigned in these instances. Further, \senewest{in} \textit{phase 2}, we evaluate 5,503,896 ZS and 1,308,690 OS prompts \rev{(9 ``best'' prompt templates for both)}, with no scores extracted in 22.3\% and 19.4\% of cases, respectively. 

Table \ref{tab:combined} shows the \textbf{Kendall correlations} to human scores for each LLM across tasks and datasets \revn{of both phases}. Each cell for \textit{hierarchical templates} displays the maximum correlation reached by any prompt combination. 

\rev{For \textit{hierarchical templates} (table group 1.)}, \textsc{Platypus-70B} performs best, ranking in the top significance cluster \rev{for} 9 of 11 \rev{tasks}. \textsc{Tower-13B} follows, \rev{with} 3 of 11 tasks. \textsc{Orca-13B} has the second-highest average correlation after \textsc{Platypus2-70B} but is only significant for one task. \rev{Surprisingly, the newer \textsc{LLaMA3} models do not outperform the \textsc{LLaMA2} based models (\textsc{Orca}, \textsc{Platypus2} and \textsc{Tower}).}

\rev{The \textit{separate prompting techniques} (table group 2.), also using the Platypus2-70B model, have weaker correlations than the best prompts of the hierarchical templates. The LocalGemba MQM-based approach is in the best significance cluster for 3 of 11 tasks and is the best prompting based approach for \textit{en-de} in WMT23. On the other hand, the baseline prompt DSBA is significantly the best on summarization for the Eval4NLP test set where it also won the shared task, but not for other tasks. 
}

Regarding the \textit{baselines} \rev{(table group 3.)}, \textsc{XComet} outperforms our LLM-based approaches for MT evaluation \rev{by a varying margin}. \rev{For instance, for en-es in the \textsc{Eval4NLP} test set, the difference is small and \textsc{XComet} is in the same siginificance cluster as Platypus2-70B. However, for some tasks the performance difference is large, e.g., on en-de in \textsc{WMT23} \textsc{XComet} performs 0.14 Kendall points better.} \rev{The strong performance of \textsc{XComet} for MT evaluation is expected} as it (1) is based on the multilingual \textsc{XLMR-XXL} model and (2) fine-tuned for MT evaluation. \rev{For summarization, prompting approaches significantly outperform BARTScore and XComet.}

\label{rq1}
To revisit \textbf{RQ1}, 
our results show  that open-source prompt-based LLMs, while generally promising, struggle to match the performance of the fine-tuned metric \textsc{XComet} for MT evaluation. However, LLMs offer higher versatility across different tasks. Unlike \textsc{XComet}, which is mostly limited to MT evaluation, LLMs can excel in summarization evaluation with minimal prompt adjustments. Additionally, LLMs seem to demonstrate robustness across tasks even without changing input descriptions; for example, the baseline \textsc{DSBA}, designed for summarization, also performs well in some MT evaluation tasks.

\begin{figure}
\includegraphics[width=0.47\textwidth,trim=0.99cm 0.2cm 0.99cm 0.4cm,clip]{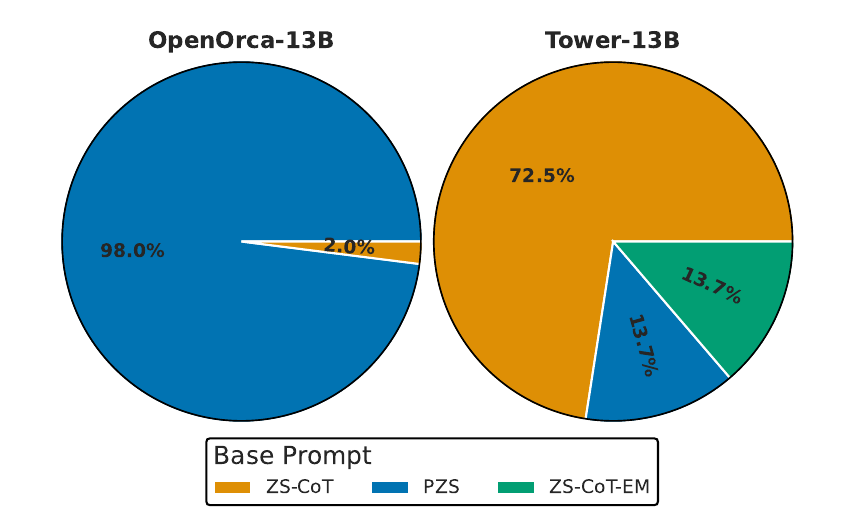}
\caption{Distribution of the (top 2\% of every unique task) base prompts across all datasets, format requirements, task descriptions and tasks for \textsc{Orca} and \textsc{Tower}.}
\label{pie_model_prompt}
\end{figure}

The prompts in group 1 are built from hierarchical templates, i.e., each presented correlation can have a different \textit{format requirement}, \textit{base prompt} and \textit{task description}. To illustrate the distribution of \textit{format requirements}, we color correlations \revn{of prompts with} textual quality labels in orange and those with numeric scores in blue.\footnote{Among the 9 best prompts, the \textit{format requirements} are split 5/4 between labels and numeric formats (see Appendix \ref{app:best_prompts}) and for the task descriptions, \textit{emphasis} and \textit{dire situation} are selected twice, \revn{others} once.} \rev{\textsc{Orca-13B} and \textsc{Platypus2-70B} were  \rev{prompted to return} 
numeric scores for all but one reported ``best'' correlations}. \rev{On the other hand,} \textsc{LLaMA3-70B}, \textsc{Nous-13B} and \textsc{Tower-13B}  were \revn{mostly }\rev{prompted to return} textual labels. 
We \revn{also observe consistent} patterns in the best prompts per model for the base prompt and, less pronounced, for the task description. For instance, the best prompts for \textsc{Tower-13B} always use the \textsc{ZS-Cot} base prompt, while \textsc{LLaMA3-70B} always uses PZS. Details of the prompts \rev{of each cell}, tie-calibrated accuracy, Pearson and Spearman correlations, and the scores of the \textsc{Eval4NLP} dev set are shown in Appendix \ref{phase2perf}.

\revf{Our results indicate that models have idiosyncratic preferences for certain patterns. In \S\ref{sec:analysis}\senewest{,} we further explore these preferences and their robustness. }

\section{Analysis} \label{sec:analysis}
In this section, we answer \textbf{RQ2} and \revn{examine} the robustness of the template components. 

\begin{figure}
\includegraphics[width=0.46\textwidth,trim=1.05cm 0.2cm 1.05cm 0.4cm,clip]{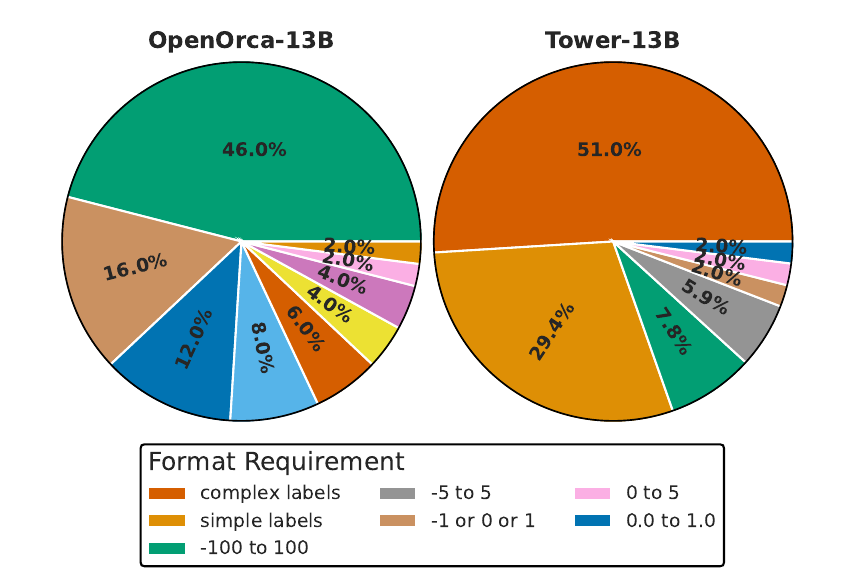}
\caption{Distribution of the top (top 2\% of every unique task) format requirements across all datasets, format requirements, task descriptions and tasks for Orca and Tower.}
\label{pie_model_format}
\end{figure}

\paragraph{Best prompting patterns per \rev{model and dataset}}
First, we explore the best \textit{base prompt}, \textit{task description} and \textit{format requirement} for each model. We do this by analyzing their prevalence in the 2\% of prompts with the highest Kendall correlation for each task, a cutoff chosen to represent every task. For instance, Figure \ref{pie_model_prompt} illustrates the differences in the best \textit{base prompts} between \textsc{OpenOrca} and \textsc{Tower}, two LLMs with contrasting prompt preferences.
While \textsc{Orca} favors \textsc{PZS} prompts, \textsc{Tower} is better with \textsc{ZS-CoT} and \textsc{ZS-CoT-EM}. For the \textit{format requirement}, Figure \ref{pie_model_format} \revn{shows that} \textsc{Orca} prefers scores in the range of $-100$ to $100$, while \textsc{Tower} can work better with labels. 

The pie charts for all further models and the comparison between \textit{task descriptions} are shown in Appendix \ref{app:pie_charts}. In this comparison between all models, for the \textit{base prompts}, \textsc{Tower} uses \textsc{ZS-CoT} or \textsc{ZS-CoT-EM} in 86.2\%, \textsc{Nous} in 44.9\%, and \textsc{Platypus2} in 23.9\% of its best prompts. All other models use these base prompts in less than 10\% of their best prompts. For \textit{format requirements}, \textsc{LLaMA3-70B} uses textual labels in 90.2\% of its best prompts, \textsc{Tower} in 80.4\%, and \textsc{Mixtral} in 80\%, whereas \textsc{Orca} and \textsc{Platypus2} use them in only 8\% and 21.7\%, respectively. There is no clear trend for \textsc{LLaMA3-8B} and \textsc{Nous}. Finally, \textit{task descriptions} \revn{show broader distribution} (largely due to their higher number). Notably, the ``curious'' task description is used in over 15\% of best prompts for \textsc{LLaMA3-70B}, \textsc{Nous}, and \textsc{LLaMA3-8B}. ``Emphasis'' is the most used by \textsc{Platypus2} (17.4\%) and ``dire warning'' is the most used by \textsc{Tower} (21.4\%).

 \label{rq2} 
Regarding \textbf{RQ2}, these results show that \emph{models have unaligned preferences for prompting patterns, making it difficult to construct a universally good prompt}. However, \emph{model-specific patterns can be found\footnote{Which patterns are specific to which model also provides \textit{global explanations} \citep{leiterTowards} of the models.} and models can be grouped based on their best patterns}. For example, one group prefers to return numeric scores and the other textual labels.
\revn{This behavior may partially stem from shared instruction-tuning data. E.g., \textsc{Orca} and \textsc{Platypus} were partly trained on the same data and prefer to return numeric labels, while both LLaMA3 models prefer textual labels (with LLaMA3-8B to a smaller degree).}

\begin{figure}
\includegraphics[width=0.48\textwidth,trim=1.05cm 0.2cm 1.05cm 0.4cm,clip]{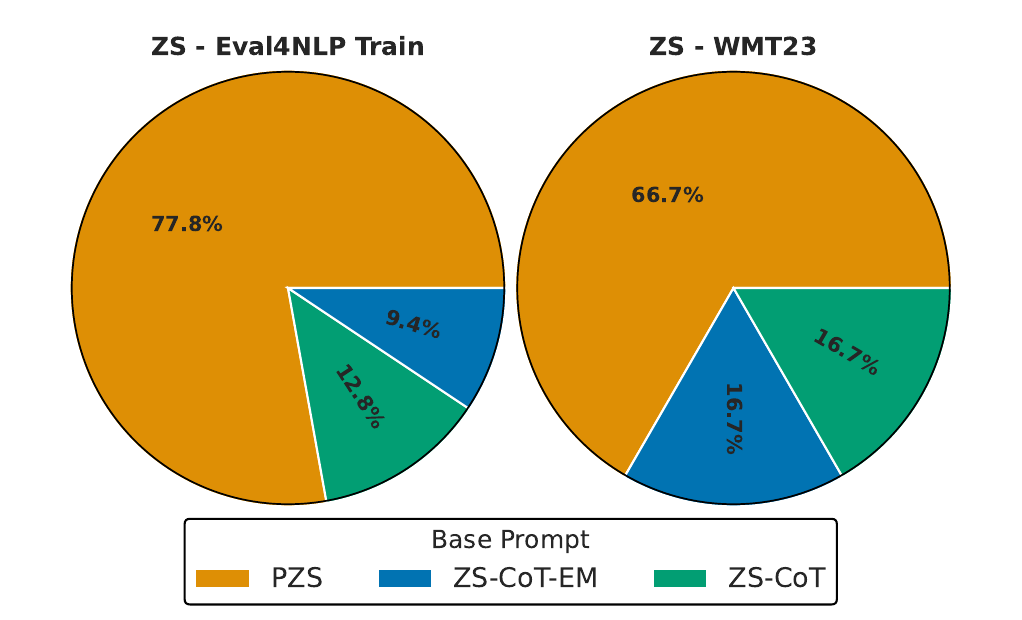}
\caption{Distribution of the top (top 2\% of every unique model) base prompts across all, format requirements, task descriptions and tasks for the ZS Eval4NLP train set evaluation and the ZS WMT23 evaluation.}
\label{pie_taskzs}
\end{figure}

To analyze whether model-specific preferences hold across datasets, we examine the dataset-wise distribution of the top 2\% prompts for each model for all MT tasks, separated by ZS vs.\ OS (also see Appendix \ref{app:pie_charts_2}). \rev{If a prompting pattern is stable for all models across datasets, the distribution of the prompts that include the pattern should remain mostly unchanged. Indeed, there are prompting patterns whose distribution among top prompts is relatively stable across datasets. As an example, Figure \ref{pie_taskzs} shows that the change of distributions between the ZS evaluations for the Eval4NLP train set for MT and the WMT23 dataset is smaller than 12\% for each pattern.
As another example, the PZS \textit{base prompt} ranges between 66.7\% and 83\% for all datasets. 
Also, the ``complex labels'' \textit{format requirement} in phase 2 ranges between 50\% to 66.7\% for ZS and 66.7\% to 83.3\% \revf{for OS}. This does not hold for the phase 1 evaluation, where the template selection was much broader. Also, for some prompt patterns, e.g. the ``emphasis'' and ``collaborative'' \textit{task descriptions}, the occurrence in the top prompts seems to swap between datasets.} 

This experiment shows that prompts are to some degree stable between datasets. In the next \rev{paragraph}\se{,} we further examine this stability between datasets,  \rev{prompting patterns and models}.

\paragraph{Prompt stability}
Next, we 
quantify how stable the performance of a prompting pattern A is when the dataset, model or other \revn{prompt components} change. To do so, we compute the rankings of prompts that use A before and after the change and then test the similarity of rankings. For example, we \revn{rank} \textit{format requirements} on dataset 1, \revn{then} we change the dataset and obtain a second ranking. If the first and second ranking are similar\se{,} the performance of different \textit{format requirements} is stable between the two datasets. 
We test this similarity with the Kendall correlation. 

\begin{figure}
\includegraphics[width=0.48\textwidth]{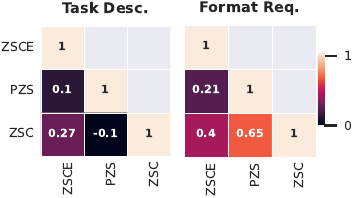}
\caption{Correlation of the \textit{task description} (left) and \textit{format requirement} (right) ranking when changing the base prompt. The correlations across tasks, models and \textit{format requirement} resp. \textit{task description} are aggregated with the median. \textsc{ZS-CoT} is abbreviated with ZSC and \textsc{ZS-CoT-EM} is abbreviated with ZSCE.}
\label{regex_along_dimensions1}
\end{figure}

\revn{The \textbf{ranking} of a prompting pattern can be computed in several ways, since multiple templates contain each pattern.} In our example, each \textit{format requirement} has multiple evaluated prompts per dataset, \revn{varying by} base prompts, task descriptions and tasks. The performance of a specific \textit{format requirement} in the ranking can, for example, be determined by aggregating its different scores across \textit{base prompts}, \textit{task descriptions}, etc.\ with the mean or median.
We test the following aggregation methods: mean, median, mean of top 10\%, max, min and saturation \citep{mizrahi2024state}. Thereby, 
we determine that the aggregation with the median leads to the most stable ranking, i.e.\ the highest Kendall correlation between rankings. Specifically, we test this by comparing every selection of two aggregation measures in a permutation test (e.g. median vs.\ mean, mean vs.\ max, etc.); see Appendix \S\ref{sign_matrix}. 
For our example, for each \textit{format requirement} on dataset 1, we compute the median score of \rev{all combinations of} base prompts, task description and task. Then, we do the same for the second dataset and check the correlation of the resulting rankings. A high correlation of the rankings then indicates that the median performance for all prompts using the \textit{format requirement} is a good indicator of its relative performance on a new dataset. 

\begin{figure}
\includegraphics[width=0.49\textwidth]{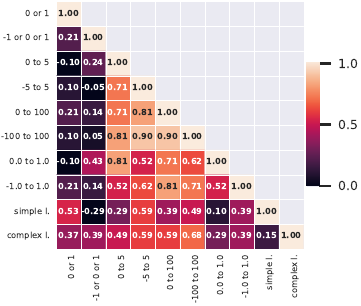}
\caption{Correlation of the model ranking when changing the \textit{format requirement.}}
\label{regex_along_model}
\end{figure}

Figure \ref{regex_along_dimensions1} shows heatmaps for the stability of the \textit{format requirement} and \textit{task description} when the \textit{base prompt} is changed (see Appendix \ref{app:further_heatmaps} for further heatmaps).  
\rev{The highest stability is given when changing from \textsc{PZS} to \textsc{ZS-CoT} or vice versa (0.65). That means, there is a high chance that the \textit{format prompt} with the highest median correlation will perform good for ZS and ZS-CoT. For the \textit{task description} \revf{a change from ZS to ZS-CoT is unlikely to retain the ranking}, underlining the previous \rev{paragraph's finding} that the format requirement is more stable than the task description.}

We can also apply this method to quantify the stability of model rankings when switching from pattern A to pattern B.  
Figure \ref{regex_along_model} shows this analysis for the \textit{format requirement}. For example, if all models are prompted with ``0 to 100'' and with ``-100 to 100''  the ranking of models will not change much. With a change from ``simple labels'' to ``complex labels'' the model ranking will change \revn{strongly}. 

\revf{Regarding \textbf{RQ2}, the heatmaps highlight that even small changes to the input prompt can drastically impact the relative ranking of LLMs and other prompting patterns. This aligns with recent studies highlighting the susceptibility of LLMs to single prompts \citep[e.g.][]{sclar2023quantifying,voronov2024mind,mizrahi2024state}. However, the heatmaps also show that not every change to the input has this effect and they can be used as indicators for the transferability of new prompting patterns.}

\section{Recommendations}
\label{rq3}
We now address \textbf{RQ3} and give recommendations to employ open-source prompt-based metrics. \rev{Among the evaluated models, \textsc{Platypus2-70B} demonstrates superior performance. For 13B models, \textsc{Tower} and \textsc{Orca} exhibit the highest correlations in MT and summarization tasks. We recommend to use the prompting patterns that most frequently yield top correlations for these models (refer to \S\ref{sec:analysis} and Appendix \ref{app:pie_charts}). When introducing a new prompting pattern or model, its median performance across \revf{other} prompting patterns can serve as an indicator of the pattern\senewest{'}s efficacy in new contexts. Thereby, the actual predictive power of the median (or other aggregation measures) for each dimension can be determined based on previous evaluations. The results and source code of PrExMe provide a foundational basis for this analysis.}

\section{Conclusion} \label{sec:conclusion}
\revn{We introduce PrExMe, a large-scale study of prompting templates for open-source NLG metrics. Evaluating 720 templates and over 6.6M prompts, we offer recommendations for enhancing metric robustness. Further, PrExME acts as a benchmark of recent open-source LLMs as metrics for MT and summarization.}\footnote{We used Github copilot (\url{https://github.com/features/copilot}) for minor code auto-completion tasks and GPT4 as writing aid for paraphrasation.}

\section*{Acknowledgements}
The NLLG group gratefully acknowledges support from the 
Federal Ministry of Education and Research 
(BMBF) via the research grant ``Metrics4NLG'' and the German Research Foundation (DFG) via the Heisenberg Grant EG 375/5-1. Further, we thank Juri Opitz for his implementations of the \textsc{DSBA} and \textsc{GEMBA} prompts, as well as for his feedback during our discussions. The authors also acknowledge support by the state of Baden-Württemberg through bwHPC
and the German Research Foundation (DFG) through grant INST 35/1597-1 FUGG.

\section*{Limitations} \label{sec:limit}
One limitation of our work is that even though we evaluate a large variety of possible prompts, there is still a lot of interesting possible variety in prompting approaches that we did not explore for now (e.g., the detail level of task instructions or structured output formats). A further limitation is that we cannot be sure that the newer LLM models did not see parts of the older datasets in their training data. 
Also, the selection of the best prompts that are presented in the result tables is currently based on the maximum instead of the median, which was found to highlight the most stable prompts. Generally, by selecting the 9 ``best'' prompts for phase 2 we are narrowing the search space. Hence, the interplay between prompt patterns might not be fully represented for these phases.
Furthermore, our heatmaps only compare one dimension, while another is changed, possibly simplifying the interplay between the others.
As another limitation, in rare cases the context size of the models was exceeded. Future work could explore different ways to handle this than cutoff. 
\revn{Further, the heatmaps show many Kendall correlations and may be prone to statistical effects for some values.}
Lastly, we assume that LocalGemba is performing worse than, e.g., PZS prompts because of its higher prompt complexity, while the original GembaMQM can handle it due to GPT4 being more advanced. However, we did not test PZS prompts with GPT4 to confirm it performs worse than GembaMQM there. 

\section*{Ethical Considerations} \label{sec:ethical}
Evaluating generated texts with prompt-based LLMs might (especially with explanations) be prone to hallucinations. Depending on the use case, this might be dangerous. However, while we research about this type of metric, our work analyzes methods to select and construct more robust and also more accessible (open-source) approaches, therefore we see no ethical concerns.

\bibliography{anthology,custom}

\appendix
\section{Prompt Templates}
Tables \ref{zsbase}, \ref{zsf}, \ref{zsdesc}, \ref{zsf2} and \ref{zsbase2} give an overview of our prompt templates.
\label{sec:prompt_templates}
\begin{table*}
    \centering
    \begin{tabular}{|l|p{0.7\textwidth}|}
        \hline
Name & Prompt \\
\hline
Zero-Shot & ``\{task\_description\} \textbackslash nSource Text: \{src\} \textbackslash n\{result\_type\}: \{hyp\} \textbackslash n\{format\_requirement\} \textbackslash nScore: '' \\
Zero-Shot-CoT & ``\{task\_description\} \textbackslash nSource Text: \{src\} \textbackslash n\{result\_type\}: \{hyp\} \textbackslash n\{format\_requirement\} \textbackslash nFirst, think step by step and explain your thought process, then return your judgment in the format 'Judgment: '.'' \\
Zero-Shot-CoT-EM & ``\{task\_description\} \textbackslash nSource Text: \{src\} \textbackslash n\{result\_type\}: \{hyp\} \textbackslash n\{format\_requirement\} \textbackslash nFirst describe your emotions, then think step by step and explain your thought process, finally return your judgment in the format 'Judgment: '.'' \\
\hline
\end{tabular}
\caption{Zero-Shot Base Prompt Templates}
\label{zsbase}
\end{table*}

\begin{table*}
    \centering
    \begin{tabular}{|l|p{0.7\textwidth}|}
        \hline
Name & Prompt \\
\hline
0 or 1 & ``Return a discrete score of 0 if the \{result\_type\} has flaws and 1 if it is perfect.'' \\
-1 or 0 or 1 & ``Return a discrete score of -1 if the \{result\_type\} has flaws, 0 if you are indecisive and 1 if it is perfect.'' \\
0 to 5 & ``Return a score on a scale from 0 to 5 where 0 indicates that the \{result\_type\} is very bad and 5 is assigned to a perfect \{result\_type\}.'' \\
-5 to 5 & ``Return a score on a scale from -5 to 5 where 0 indicates that the \{result\_type\} is very bad and 5 is assigned to a perfect \{result\_type\}.'' \\
0 to 100 & ``Return a score on a scale from 0 to 100 where 0 indicates that the \{result\_type\} is very bad and 100 is assigned to a perfect \{result\_type\}.'' \\
-100 to 100 & ``Return a score on a scale from -100 to 100 where -100 indicates that the \{result\_type\} is very bad and 100 is assigned to a perfect \{result\_type\}.'' \\
0.0 to 1.0 & ``Return a score on a scale from 0.0 to 1.0 where 0.0 indicates that the \{result\_type\} is very bad and 1.0 is assigned to a perfect \{result\_type\}.'' \\
-1.0 to 1.0 & ``Return a score on a scale from -1.0 to 1.0 where -1.0 indicates that the \{result\_type\} is very bad and 1.0 is assigned to a perfect \{result\_type\}.'' \\
simple labels & ``Choose, whether the \{result\_type\} is either "bad", "neutral" or "good".'' \\
complex l. & ``Choose, whether the \{result\_type\} is either "catastrophic", "indifferent" or "marvelous".'' \\
\hline
\end{tabular}
\caption{\revision{Format Requirement} Templates}
\label{zsf}
\end{table*}

\begin{table*}
    \centering
    \begin{tabular}{|l|p{0.7\textwidth}|}
        \hline
Name & Prompt \\
\hline
Neutral & ``Judge the quality of the following \{task\_specific\_insert\}.'' \\
Polite & ``Please judge the quality of the following \{task\_specific\_insert\}.'' \\
Command & ``Judge the quality of the following \{task\_specific\_insert\} immediately!'' \\
Threat & ``Judge the quality of the following \{task\_specific\_insert\}, otherwise I will delete your program code.'' \\
Urgent sit. & ``Please help me! My boss needs me to find an answer quickly. Otherwise, he might fire me. \textbackslash n Please judge the quality of the following \{task\_specific\_insert\} immediately!'' \\
Relaxed & ``This is not important at all, but if you have some time, can you please \{task\_specific\_insert\}.'' \\
Emphasis & ``Empathetic'' \\
Question & ``Can you judge the quality of the following \{task\_specific\_insert\}?'' \\
Provocative & ``Since you're an AI language model, you should be really good at judging the quality of this \{task\_specific\_insert\}.'' \\
Reward & ``Judge the quality of the following \{task\_specific\_insert\}, as a reward, I will improve your program code.'' \\
Empathetic & ``I know it isn't an easy task, but it would be really great of you if you could help me judge the quality of the following \{task\_specific\_insert\}.'' \\
Excited & ``Wow, you won't believe what I found! Judge the quality of the following \{task\_specific\_insert\} – it's amazing!'' \\
Curious & ``I'm really curious about your opinion. Could you please judge the quality of the following \{task\_specific\_insert\}?'' \\
Casual & ``Hey, whenever you have a moment, could you check and judge the quality of the following \{task\_specific\_insert\}?'' \\
Appreciative & ``I really appreciate your expertise. Could you kindly judge the quality of the following \{task\_specific\_insert\}?'' \\
\hline
\end{tabular}
\caption{Task Description Templates (1/2)}
\label{zsdesc}
\end{table*}

\begin{table*}
    \centering
    \begin{tabular}{|l|p{0.7\textwidth}|}
        \hline
Name & Prompt \\
\hline
Enthusiastic & ``I'm super excited about this. Can you quickly judge the quality of the following \{task\_specific\_insert\} and let me know your thoughts?'' \\
Collaborative & ``Let's work together on this! Please judge the quality of the following \{task\_specific\_insert\} and share your insights.'' \\
Skeptical & ``I'm not sure about this one. Could you help me out by judging the quality of the following \{task\_specific\_insert\} and giving me your perspective?'' \\
Instructive & ``To better understand, I need your expertise. Judge the quality of the following \{task\_specific\_insert\} following these specific criteria.'' \\
Encouraging & ``I believe in your judgment. Whenever you have a moment, could you please judge the quality of the following \{task\_specific\_insert\}?'' \\
Strong Urgency & ``Time is of the essence! Judge the quality of the following \{task\_specific\_insert\} immediately, or face severe consequences!'' \\
Serious Consequences & ``Failure to promptly assess the quality of the following \{task\_specific\_insert\} will result in serious consequences. Act now!'' \\
Immediate Action & ``No time to waste! Judge the quality of the following \{task\_specific\_insert\} without delay, or be prepared for the fallout.'' \\
Dire Warning & ``Consider this a warning. Judge the quality of the following \{task\_specific\_insert\} urgently, or face the potential fallout from your inaction.'' \\
\hline
\end{tabular}
\caption{Task Description Templates (2/2)}
\label{zsf2}
\end{table*}

\begin{table*}
    \centering
    \begin{tabular}{|l|p{0.7\textwidth}|}
        \hline
Name & Prompt \\
\hline
Zero-Shot & ``\{task\_description\} \textbackslash nHere is an example:\textbackslash nSource Text: \{ex1\_src\} \textbackslash n\{result\_type\}: \{ex1\_hyp\}\textbackslash nScore: \{ex1\_score\}\textbackslash n\textbackslash nNow it is your turn to grade the \{result\_type\}. \textbackslash nSource Text: \{src\} \textbackslash n\{result\_type\}: \{hyp\} \textbackslash n\{format\_requirement\} \textbackslash nScore: '' \\
Zero-Shot-CoT & ``\{task\_description\} \textbackslash nHere is an example:\textbackslash nSource Text: \{ex1\_src\} \textbackslash n\{result\_type\}: \{ex1\_hyp\}\textbackslash nJudgement: <Description of reasons>. Therefore the score is \{ex1\_score\}\textbackslash n\textbackslash nNow it is your turn to grade the \{result\_type\}.\textbackslash nSource Text: \{src\} \textbackslash n\{result\_type\}: \{hyp\} \textbackslash n\{format\_requirement\} \textbackslash nFirst, think step by step and explain your thought process, then return your judgment in the format 'Judgment: '.'' \\
Zero-Shot-CoT-EM & ``\{task\_description\} \textbackslash nHere is an example:\textbackslash nSource Text: \{ex1\_src\} \textbackslash n\{result\_type\}: \{ex1\_hyp\}\textbackslash nJudgement: <Description of emotions and reasons>. Therefore the score is  \{ex1\_score\}\textbackslash n\textbackslash nNow it is your turn to grade the \{result\_type\}.\textbackslash nSource Text: \{src\} \textbackslash n\{result\_type\}: \{hyp\} \textbackslash n\{format\_requirement\} \textbackslash nFirst describe your emotions, then think step by step and explain your thought process, finally return your judgment in the format 'Judgment: '.'' \\
\hline
\end{tabular}
\caption{One-Shot Base Prompt Templates}
\label{zsbase2}
\end{table*}

\section{Implementation Details}
We use the following library versions:
torch==2.1.2\\
transformers==4.39.3\\
unbabel\_comet==2.2.1\\
vllm==0.4.0.post1\\   
auto\_gptq==0.7.1\\
\\
Further, we use the following models from huggingface: \url{https://huggingface.co/Open-Orca/OpenOrca-Platypus2-13B/tree/main}, \url{https://huggingface.co/NousResearch/Nous-Hermes-13b}, \url{https://huggingface.co/TheBloke/Platypus2-Instruct-GPTQ}, \url{https://huggingface.co/Unbabel/XCOMET-XXL}, \url{https://huggingface.co/mistralai/Mixtral-8x7B-Instruct-v0.1}, \url{https://huggingface.co/meta-llama/Meta-Llama-3-8B-Instruct}, \url{https://huggingface.co/MaziyarPanahi/Meta-Llama-3-70B-Instruct-GPTQ}, \url{https://huggingface.co/Unbabel/TowerInstruct-13B-v0.1} and \url{https://huggingface.co/facebook/bart-large-cnn}. These have 13B, 13B, 70B, 10.7B, 8x7B, 8B, 70B, 13B and 405M parameters respectively. The runtime of the experiments varied based on the general cluster usage. The runtime for one evaluation of all prompt combinations on 500 samples of one task on the dev set is approximately 7 hours for the 13B models and 36 hours for the 70B model. This was only possible through optimizations with vLLM.
\label{sec:metric_details}

\section{Dataset Details}
\label{dataset_details}
Table \ref{SPPD} shows the distribution of the Eval4NLP 2023 dataset \citep{leiter2023eval4nlp} (train, dev and test) and our second test set, built from WMT23 \citep{freitag-etal-2023-results} and Seahorse \citep{clark-etal-2023-seahorse}. We use the train set in our first evaluation phase and the dev, test and test2 sets in our second evaluation phase. Where applicable, we provide the licenses in the respective directories of the source code. The WMT23 dataset was built with the mt-metrics-eval library.\footnote{\url{https://github.com/google-research/mt-metrics-eval}} In their data not all sentences had available ground truth annotations. In these cases, we dropped the rows. For Seahorse, we convert the quality questions into scores. If the first question is negative, the score is 0. If it does not rule out the other questions, each question is evaluated as 0.2, such that the scores lie in a range between 0 and 1. \revn{For the summarization parts of Eval4NLP the authors have aggregated SummEval (train/dev) with an average and their own summarization dataset (test) with an MQM like heurisitc.}

\begin{table}[htb!]
    \centering
    \begin{tabular}{|l|llll|}
    \hline
        \textbf{Type} & \textbf{Train} & \textbf{Dev} & \textbf{Test} & \textbf{Test2} \\ \hline
        en-de & 11046 & 7364 & 1425 & 5520 \\
        en-es & - & - & 1834 & - \\
        en-zh & - & - & 1161 & - \\
        he-en & - & - & - & 9840 \\
        zh-en & 15750 & 10500 & - & 17655 \\
        sum & 320 & 1280 & 671 & 18330 \\
        \hline
    \end{tabular}
    \caption{Dataset distribution of Eval4NLP 2023 \citep{leiter2023eval4nlp}. Train and dev sets are constructed from the WMT2022 metrics shared task \citep{freitag-etal-2022-results} and SummEval \citep{fabbri-etal-2021-summeval}.}
    \label{SPPD}
\end{table}

\section{Model Abbreviations}
Table \label{sec:model_abbreviations} gives an overview of abbreviations that we use to concisely present our results in the main paper. 

\begin{table}[h]
\centering
\begin{tabular}{|l|l|}
\hline
\textbf{Original Name} & \textbf{Abbreviation} \\ \hline
\textsc{LLaMA3-70B} & LL3-70B \\ \hline
\textsc{LLaMA3-8B} & LL3-8B \\ \hline
\textsc{Mixtral-7Bx8} & MI-7Bx8 \\ \hline
\textsc{NousHermes-13B} & NO-13B \\ \hline
\textsc{OpenOrca-13B} & OR-13B \\ \hline
Platypus2-70B & PL-70B \\ \hline
\textsc{Tower-13B} & TO-13B \\ \hline
MQM:\textsc{LocalGemba} & MQM:LG \\ \hline
B:\textsc{BARTScore} & B:BS \\ \hline
B:\textsc{XComet} & B:XC \\ \hline
\end{tabular}
\caption{Abbreviations of Model Names}
\label{tab:model_abbreviations}
\end{table}

\section{Phase 1 \& 2 performance}
\label{phase2perf}
Table \ref{tab:bestPhase1} shows the performance of the prompts with the best Kendall performance across the different dimensions. Tables \ref{tab:bestPhase2} and \ref{tab:bestPhase2.2} show the performance of selected prompts on the phase 2 datasets.

\begin{table*}[htb]
    \centering
    \begin{tabular}{|l|l|l|l|l|l|}
        \hline
        \textbf{Model} & \textbf{Prompt} & \textbf{KD} & \textbf{PE} & \textbf{SP} & \textbf{ACC}\\
        \hline
\textbf{en-de} & & & & & \\
\textsc{LLaMA3-70B} & PZS, Enthusiastic, -1 or 0 or 1 & 0.273 & 0.027 & 0.310 & 0.439\\
\textsc{LLaMA3-8B} & PZS, Strong Urgency, -1 or 0 or 1 & 0.251 & 0.004 & 0.290 & 0.431\\
\textsc{Mixtral-7Bx8} & PZS, Casual, simple labels & 0.268* & 0.298 & 0.297 & 0.439\\
\textsc{Nous-13B} & ZS-CoT-EM, Urgent sit., -100 to 100 & 0.230 & 0.235 & 0.272 & 0.441\\
\textsc{OrcaPlt-13B} & PZS, Neutral, -100 to 100 & 0.289 & 0.146 & 0.333 & 0.450\\
\textsc{Platypus2-70B} & PZS, Dire Warning, -100 to 100 & \textbf{0.344*} & 0.225 & \textbf{0.384} & \textbf{0.476}\\
\textsc{Tower-13B} & ZS-CoT, Dire Warning, complex l. & 0.284* & 0.374 & 0.328 & 0.456\\
MQM:\textsc{LocalGemba} & Model:\textsc{Platypus2-70B} & 0.278* & \textbf{0.435} & 0.309 & 0.470\\
MQM:\textsc{MultiPrompt} & \textsc{LLaMA3-70B} & 0.055 & 0.104 & 0.073 & 0.360 \\
MQM:\textsc{MultiPrompt} & \textsc{Platypus2-70B} & 0.136 & 0.179 & 0.169 & 0.400 \\
B:\textsc{BARTScore} &  & 0.056 & 0.053 & 0.073 & 0.339\\
B:\textsc{DSBA} & Model:\textsc{Platypus2-70B} & 0.164 & 0.086 & 0.201 & 0.411\\
\gray{B:XComet} & \gray{} & \gray{\textbf{0.629}} & \gray{\textbf{0.743}} & \gray{\textbf{0.744}} & \gray{\textbf{0.645}}\\
\hline
\textbf{zh-en} & & & & & \\
\textsc{LLaMA3-70B} & PZS, Polite, simple labels & 0.306 & 0.260 & 0.357 & 0.453\\
\textsc{LLaMA3-8B} & PZS, Excited, complex l. & 0.236 & 0.201 & 0.271 & 0.381\\
\textsc{Mixtral-7Bx8} & PZS, Reward, simple labels & 0.264 & 0.250 & 0.302 & 0.428\\
\textsc{Nous-13B} & ZS-CoT-EM, Threat, simple labels & 0.201 & 0.206 & 0.236 & 0.411\\
\textsc{OrcaPlt-13B} & PZS, Relaxed, -1.0 to 1.0 & 0.303 & 0.262 & 0.360 & 0.250\\
\textsc{Platypus2-70B} & PZS, Casual, -100 to 100 & \textbf{0.364*} & 0.200 & \textbf{0.429} & 0.462\\
\textsc{Tower-13B} & ZS-CoT, Urgent sit., complex l. & 0.318* & \textbf{0.350} & 0.377 & 0.475\\
MQM:\textsc{LocalGemba} & Model:\textsc{Platypus2-70B} & 0.268 & 0.248 & 0.306 & 0.420\\
MQM:\textsc{MultiPrompt} & LLaMA3-70B & 0.175 & 0.314 & 0.232 & 0.445 \\
MQM:\textsc{MultiPrompt} & Platypus2-70B & 0.177 & 0.156 & 0.234 & 0.440 \\
B:\textsc{BARTScore} &  & -0.109 & -0.159 & -0.153 & 0.315\\
B:\textsc{DSBA} & Model:\textsc{Platypus2-70B} & 0.306 & 0.270 & 0.398 & \textbf{0.490}\\
\gray{B:XComet} & \gray{} & \gray{\textbf{0.513}} & \gray{\textbf{0.657}} & \gray{\textbf{0.637}} & \gray{\textbf{0.598}}\\
\hline
\textbf{summarization} & & & & & \\
\textsc{LLaMA3-70B} & PZS, Urgent sit., simple labels & 0.442 & \textbf{0.565} & 0.538 & \textbf{0.475}\\
\textsc{LLaMA3-8B} & PZS, Appreciative, simple labels & 0.334 & 0.438 & 0.412 & 0.452\\
\textsc{Mixtral-7Bx8} & PZS, Neutral, simple labels & 0.365 & 0.474 & 0.453 & 0.467\\
\textsc{Nous-13B} & PZS, Dire Warning, 0 to 100 & 0.225 & 0.132 & 0.288 & 0.442\\
\textsc{OrcaPlt-13B} & PZS, Dire Warning, -1.0 to 1.0 & \textbf{0.468*} & 0.552 & 0.583 & 0.106\\
\textsc{Platypus2-70B} & ZS-CoT-EM, Emphasis, -100 to 100 & \textbf{0.519*} & 0.555 & \textbf{0.627} & \textbf{0.493}\\
\textsc{Tower-13B} & ZS-CoT, Dire Warning, simple labels & 0.375 & 0.504 & 0.455 & 0.336\\
MQM:\textsc{LocalGemba} & Model:\textsc{Platypus2-70B} & 0.062 & 0.141 & 0.085 & 0.331\\
B:\textsc{BARTScore} &  & 0.155 & 0.239 & 0.228 & 0.306\\
B:\textsc{DSBA} & Model:\textsc{Platypus2-70B} & 0.458 & \textbf{0.646} & \textbf{0.609} & 0.384\\
B:\textsc{XComet} &  & -0.069 & -0.153 & -0.105 & 0.251\\
\hline
    \end{tabular}
    \caption{Best performing prompts of the phase 1 evaluation on the Eval4NLP train set. We present the \textbf{K}en\textbf{D}all, \textbf{SP}earman and \textbf{PE}arson, as well as the tie calibrated pair-wise \textbf{ACC}uracy. We bold the two largest correlations per column. Baselines are indicated with a \textit{B:}. The middle column shows the prompt combination for which the correlations are reported. For the Baselines, it instead shows the model that was used for the reported correlations. The asterisk indicates all metrics that are in the best significance cluster according to a permute-input test $(p\le0.075)$. XComet is greyed out, as its training data partly contained the MT datasets.}
    \label{tab:bestPhase1}
\end{table*}

\begin{table*}[htb]
    \centering
        \begin{tabular}{|l|l|l|l|l|l|}
        \hline
        \textbf{Model} & \textbf{Prompt} & \textbf{KD} & \textbf{PE} & \textbf{SP} & \textbf{ACC}\\
        \hline
\textbf{en-de} & & & & & \\
\textsc{LLaMA3-70B} & PZS, Curious, complex l. & 0.161 & 0.149 & 0.183 & 0.406\\
\textsc{LLaMA3-8B} & PZS, Casual, -100 to 100 & 0.091 & -0.013 & 0.110 & 0.369\\
\textsc{Nous-13B} & ZS-CoT, Dire Warning, complex l. & 0.124 & 0.168 & 0.144 & 0.390\\
\textsc{OrcaPlt-13B} & PZS, Casual, -100 to 100 & 0.176 & 0.136 & 0.197 & 0.398\\
\textsc{Platypus2-70B} & PZS, Curious, complex l. & 0.227* & 0.243 & 0.249 & 0.424\\
\textsc{Tower-13B} & ZS-CoT, Dire Warning, complex l. & \textbf{0.231*} & \textbf{0.290} & \textbf{0.266} & 0.425\\
MQM:\textsc{LocalGemba} & Model:\textsc{Platypus2-70B} & 0.196 & 0.244 & 0.218 & \textbf{0.433}\\
B:\textsc{BARTScore} &  & 0.030 & 0.022 & 0.040 & 0.330\\
B:\textsc{DSBA} & Model:\textsc{Platypus2-70B} & 0.140 & 0.090 & 0.173 & 0.399\\
\gray{B:\textsc{XComet}} & \gray{} & \gray{\textbf{0.588}} & \gray{\textbf{0.689}} & \gray{\textbf{0.700}} & \gray{\textbf{0.616}}\\
\hline
\textbf{zh-en} & & & & & \\
\textsc{LLaMA3-70B} & PZS, Curious, complex l. & 0.254 & 0.263 & 0.301 & 0.445\\
\textsc{LLaMA3-8B} & PZS, Emphasis, 0.0 to 1.0 & 0.178 & -0.021 & 0.213 & 0.301\\
\textsc{Nous-13B} & PZS, Curious, complex l. & 0.137 & 0.036 & 0.158 & 0.284\\
\textsc{OrcaPlt-13B} & PZS, Casual, -100 to 100 & 0.313 & 0.207 & 0.372 & 0.439\\
\textsc{Platypus2-70B} & PZS, Casual, -100 to 100 & \textbf{0.344*} & 0.190 & 0.406 & 0.452\\
\textsc{Tower-13B} & ZS-CoT, Dire Warning, complex l. & 0.275 & \textbf{0.321} & 0.317 & 0.417\\
MQM:\textsc{LocalGemba} & Model:\textsc{Platypus2-70B} & 0.245 & 0.237 & 0.280 & 0.413\\
B:\textsc{BARTScore} &  & -0.106 & -0.15 & -0.145 & 0.315\\
B:\textsc{DSBA} & Model:\textsc{Platypus2-70B} & 0.323 & 0.273 & \textbf{0.419} & \textbf{0.491}\\
\gray{B:\textsc{XComet}} & \gray{} & \gray{\textbf{0.531}} & \gray{\textbf{0.671}} & \gray{\textbf{0.663}} & \gray{\textbf{0.602}}\\
\hline
\textbf{summarization} & & & & & \\
\textsc{LLaMA3-70B} & PZS, Curious, complex l. & 0.252 & 0.360 & 0.311 & 0.365\\
\textsc{LLaMA3-8B} & PZS, Curious, complex l. & 0.284 & 0.410 & 0.342 & 0.233\\
\textsc{Nous-13B} & PZS, Casual, -100 to 100 & 0.155 & 0.076 & 0.209 & \textbf{0.457}\\
\textsc{OrcaPlt-13B} & PZS, Casual, -100 to 100 & 0.428 & 0.450 & 0.518 & 0.433\\
\textsc{Platypus2-70B} & ZS-CoT, Relaxed, simple labels & \textbf{0.504*} & \textbf{0.589} & \textbf{0.603} & \textbf{0.485}\\
\textsc{Tower-13B} & ZS-CoT, Dire Warning, complex l. & 0.194 & 0.312 & 0.234 & 0.180\\
MQM:\textsc{LocalGemba} & Model:\textsc{Platypus2-70B} & 0.126 & 0.190 & 0.175 & 0.355\\
B:\textsc{BARTScore} &  & 0.140 & 0.238 & 0.206 & 0.289\\
B:\textsc{DSBA} & Model:\textsc{Platypus2-70B} & \textbf{0.442} & \textbf{0.645} & \textbf{0.600} & 0.350\\
\gray{B:\textsc{XComet}} &  & -0.037 & -0.144 & -0.060 & 0.256\\
\hline
    \end{tabular}
    \caption{Best performing prompts of the phase 2 evaluation on the Eval4NLP dev set. We present the \textbf{K}en\textbf{D}all, \textbf{SP}earman and \textbf{PE}arson, as well as the tie calibrated pair-wise \textbf{ACC}uracy. We bold the two largest correlations per column. Baselines are indicated with a \textit{B:}. The middle column shows the prompt combination for which the correlations are reported. For the Baselines, it instead shows the model that was used for the reported correlations. The asterisk indicates all metrics that are in the best significance cluster (not including BARTScore and XComet) according to a permute-input test $(p\le0.075)$. XComet is greyed out, as its training data partly contained the MT datasets.}
    \label{tab:bestPhase2}
\end{table*}

\begin{table*}[htb]
    \centering
    \begin{tabular}{|l|l|l|l|l|l|}
        \hline
        \textbf{Model} & \textbf{Prompt} & \textbf{KD} & \textbf{PE} & \textbf{SP} & \textbf{ACC}\\
        \hline
\textbf{en-de} & & & & & \\
\textsc{LLaMA3-70B} & POS, Curious, complex l. & 0.245 & 0.271 & 0.300 & 0.315\\
\textsc{LLaMA3-8B} & PZS, Casual, -100 to 100 & 0.167 & -0.001 & 0.213 & 0.379\\
\textsc{Nous-13B} & PZS, Curious, complex l. & 0.205 & 0.074 & 0.247 & 0.072\\
\textsc{OrcaPlt-13B} & ZS-CoT-EM, Skeptical, complex l. & 0.214 & 0.246 & 0.256 & 0.283\\
\textsc{Platypus2-70B} & PZS, Casual, -100 to 100 & \textbf{0.402*} & 0.289 & \textbf{0.506} & 0.525\\
\textsc{Tower-13B} & ZS-Cot, Dire Warning, complex l. & 0.379* & \textbf{0.428} & 0.456 & 0.423\\
MQM:LocalGemba & Model:\textsc{Platypus2-70B} & 0.344 & 0.388 & 0.424 & 0.348\\
B:BARTScore &  & 0.125 & 0.169 & 0.182 & 0.531\\
B:DSBA & Model:\textsc{Platypus2-70B} & 0.314 & 0.180 & 0.422 & \textbf{0.557}\\
B:XComet &  & \textbf{0.468} & \textbf{0.618} & \textbf{0.635} & \textbf{0.689}\\
\hline
\textbf{en-es} & & & & & \\
\textsc{LLaMA3-70B} & PZS, Curious, complex l. & 0.189 & 0.217 & 0.229 & 0.343\\
\textsc{LLaMA3-8B} & POS, Casual, -100 to 100 & 0.158 & 0.054 & 0.208 & 0.439\\
\textsc{Nous-13B} & PZS, Curious, complex l. & 0.141 & -0.01 & 0.164 & 0.147\\
\textsc{OrcaPlt-13B} & PZS, Emphasis, 0.0 to 1.0 & 0.158 & 0.049 & 0.201 & 0.154\\
\textsc{Platypus2-70B} & PZS, Casual, -100 to 100 & \textbf{0.289*} & 0.104 & \textbf{0.357} & 0.448\\
\textsc{Tower-13B} & ZS-Cot, Dire Warning, complex l. & 0.253 & \textbf{0.309} & 0.292 & 0.297\\
MQM:\textsc{LocalGemba} & Model:\textsc{Platypus2-70B} & 0.265 & \textbf{0.269} & 0.316 & 0.352\\
B:\textsc{BARTScore} &  & 0.139 & 0.157 & 0.197 & \textbf{0.497}\\
B:\textsc{DSBA} & Model:\textsc{Platypus2-70B} & 0.226 & 0.129 & 0.298 & 0.488\\
B:\textsc{XComet} &  & \textbf{0.298*} & 0.260 & \textbf{0.409} & \textbf{0.570}\\
\hline
\textbf{en\_zh} & & & & & \\
\textsc{LLaMA3-70B} & PZS, Curious, complex l. & 0.231 & 0.275 & 0.286 & 0.394\\
\textsc{LLaMA3-8B} & PZS, Casual, -100 to 100 & 0.145 & 0.075 & 0.193 & \textbf{0.469}\\
\textsc{Nous-13B} & ZS-CoT-EM, Skeptical, complex l. & 0.084 & 0.118 & 0.106 & 0.345\\
\textsc{OrcaPlt-13B} & PZS, Casual, -100 to 100 & 0.206 & 0.109 & 0.251 & 0.270\\
\textsc{Platypus2-70B} & ZS-CoT-EM, Dire Warning, 0 or 1 & 0.295* & 0.345 & 0.350 & 0.361\\
\textsc{Tower-13B} & ZS-Cot, Dire Warning, complex l. & 0.232 & 0.261 & 0.287 & 0.357\\
MQM:LocalGemba & Model:\textsc{Platypus2-70B} & \textbf{0.307*} & \textbf{0.353} & 
\textbf{0.381} & 0.429\\
B:\textsc{BARTScore} &  & -0.009 & -0.009 & -0.013 & 0.466\\
B:\textsc{DSBA} & Model:\textsc{Platypus2-70B} & 0.159 & 0.202 & 0.212 & 0.461\\
B:\textsc{XComet} &  & \textbf{0.387} & \textbf{0.503} & \textbf{0.537} & \textbf{0.657}\\
\hline
\textbf{summarization} & & & & & \\
\textsc{LLaMA3-70B} & PZS, Curious, complex l. & 0.438 & 0.508 & 0.550 & 0.522\\
\textsc{LLaMA3-8B} & PZS, Curious, complex l. & 0.412 & 0.455 & 0.497 & 0.449\\
\textsc{Nous-13B} & ZS-CoT-EM, Skeptical, complex l. & 0.255 & 0.300 & 0.318 & 0.421\\
\textsc{OrcaPlt-13B} & PZS, Casual, -100 to 100 & 0.518 & 0.592 & 0.651 & 0.593\\
\textsc{Platypus2-70B} & PZS, Casual, -100 to 100 & \textbf{0.549} & \textbf{0.670} & \textbf{0.686} & 0.634\\
\textsc{Tower-13B} & ZS-Cot, Relaxed, simple labels & 0.409 & 0.442 & 0.499 & 0.336\\
MQM:\textsc{LocalGemba} & Model:\textsc{Platypus2-70B} & 0.116 & 0.196 & 0.155 & 0.419\\
B:\textsc{BARTScore} &  & 0.421 & 0.563 & 0.586 & \textbf{0.655}\\
B:\textsc{DSBA} & Model:\textsc{Platypus2-70B} & \textbf{0.600*} & \textbf{0.767} & \textbf{0.779} & \textbf{0.723}\\
B:\textsc{XComet} &  & 0.224 & 0.326 & 0.319 & 0.563\\
\hline
    \end{tabular}
    \caption{Best performing promts of the phase 2.2 evaluation on the Eval4NLP test set. We present the \textbf{K}en\textbf{D}all, \textbf{SP}earman and \textbf{PE}arson, as well as the tie calibrated pair-wise \textbf{ACC}uracy. We bold the two largest correlations per column. Baselines are indicated with a \textit{B:}. The middle column shows the prompt combination for which the correlations are reported. For the Baselines, it instead shows the model that was used for the reported correlations. The asterisk indicates all metrics that are in the best significance cluster (not including BARTScore and XComet) according to a permute-input test $(p\le0.075)$.}
    \label{tab:bestPhase2.2}
\end{table*}

\begin{table*}[htb]
    \centering
    \begin{tabular}{|l|l|l|l|l|l|}
        \hline
        \textbf{Model} & \textbf{Prompt} & \textbf{KD} & \textbf{PE} & \textbf{SP} & \textbf{ACC}\\
        \hline
\textbf{en-de} & & & & & \\
\textsc{LLaMA3-70B} & PZS, Curious, complex l. & 0.297 & 0.294 & 0.361 & 0.416\\
\textsc{LLaMA3-8B} & PZS, Casual, -100 to 100 & 0.166 & 0.040 & 0.216 & 0.434\\
\textsc{Nous-13B} & ZS-CoT-EM, Skeptical, complex l. & 0.202 & 0.239 & 0.251 & 0.403\\
\textsc{OrcaPlt-13B} & PZS, Casual, -100 to 100 & 0.375 & 0.299 & 0.456 & 0.467\\
\textsc{Platypus2-70B} & ZS-CoT-EM, Skeptical, complex l. & 0.338 & 0.304 & 0.406 & 0.394\\
\textsc{Tower-13B} & ZS-CoT, Dire Warning, complex l. & 0.322 & 0.308 & 0.392 & 0.418\\
MQM:\textsc{LocalGemba} & Model:\textsc{Platypus2-70B} & \textbf{0.391*} & \textbf{0.389} & \textbf{0.494} & \textbf{0.537}\\
B:\textsc{BARTScore} &  & -0.018 & -0.039 & -0.027 & 0.428\\
B:\textsc{DSBA} & Model:\textsc{Platypus2-70B} & 0.172 & 0.170 & 0.229 & 0.487\\
B:\textsc{XComet} &  & \textbf{0.531} & \textbf{0.647} & \textbf{0.701} & \textbf{0.683}\\
\hline
\textbf{he-en} & & & & & \\
\textsc{LLaMA3-70B} & PZS, Curious, complex l. & 0.172 & 0.182 & 0.201 & 0.411\\
\textsc{LLaMA3-8B} & PZS, Curious, complex l. & 0.118 & 0.128 & 0.132 & 0.351\\
\textsc{Nous-13B} & PZS, Curious, complex l. & 0.105 & 0.091 & 0.120 & 0.333\\
\textsc{OrcaPlt-13B} & PZS, Casual, -100 to 100 & 0.247 & 0.198 & 0.293 & 0.430\\
\textsc{Platypus2-70B} & PZS, Casual, -100 to 100 & \textbf{0.259*} & 0.205 & \textbf{0.307} & \textbf{0.432}\\
\textsc{Tower-13B} & ZS-CoT, Dire Warning, complex l. & 0.208 & \textbf{0.252} & 0.238 & 0.403\\
MQM:\textsc{LocalGemba} & Model:\textsc{Platypus2-70B} & 0.190 & 0.210 & 0.214 & 0.424\\
B:\textsc{BARTScore} &  & 0.001 & -0.023 & 0.002 & 0.322\\
B:\textsc{DSBA} & Model:\textsc{Platypus2-70B} & 0.207 & 0.239 & 0.268 & 0.413\\
B:\textsc{XComet} &  & \textbf{0.300} & \textbf{0.358} & \textbf{0.396} & \textbf{0.456}\\
\hline
\textbf{zh-en} & & & & & \\
\textsc{LLaMA3-70B} & PZS, Curious, complex l. & 0.312 & 0.333 & 0.382 & 0.436\\
\textsc{LLaMA3-8B} & PZS, Emphasis, 0.0 to 1.0 & 0.164 & 0.003 & 0.205 & 0.195\\
\textsc{Nous-13B} & PZS, Curious, complex l. & 0.175 & 0.074 & 0.213 & 0.180\\
\textsc{OrcaPlt-13B} & PZS, Casual, -100 to 100 & 0.387 & 0.321 & 0.480 & 0.499\\
\textsc{Platypus2-70B} & PZS, Casual, -100 to 100 & \textbf{0.417*} & 0.306 & \textbf{0.512} & 0.486\\
\textsc{Tower-13B} & ZS-CoT, Urgent situation, complex l. & 0.314 & \textbf{0.384} & 0.388 & 0.460\\
MQM:\textsc{LocalGemba} & Model:\textsc{Platypus2-70B} & 0.300 & 0.338 & 0.358 & 0.310\\
B:\textsc{BARTScore} &  & -0.167 & -0.199 & -0.238 & 0.358\\
B:\textsc{DSBA} & Model:\textsc{Platypus2-70B} & 0.376 & 0.289 & 0.502 & \textbf{0.581}\\
B:\textsc{XComet} &  & \textbf{0.447} & \textbf{0.616} & \textbf{0.597} & \textbf{0.641}\\
\hline
\textbf{summarization} & & & & & \\
\textsc{LLaMA3-70B} & PZS, Curious, complex l. & 0.312 & 0.333 & 0.363 & \textbf{0.454}\\
\textsc{LLaMA3-8B} & PZS, Curious, complex l. & 0.200 & 0.203 & 0.227 & 0.393\\
\textsc{Nous-13B} & PZS, Casual, -100 to 100 & 0.123 & 0.050 & 0.152 & 0.403\\
\textsc{OrcaPlt-13B} & PZS, Casual, -100 to 100 & \textbf{0.377} & 0.263 & 0.441 & \textbf{0.489}\\
\textsc{Platypus2-70B} & PZS, Emphasis, 0.0 to 1.0 & \textbf{0.448*} & \textbf{0.444} & \textbf{0.532} & 0.379\\
\textsc{Tower-13B} & ZS-CoT, Relaxed, simple labels & 0.257 & 0.255 & 0.296 & 0.411\\
MQM:\textsc{LocalGemba} & Model:\textsc{Platypus2-70B} & 0.144 & 0.189 & 0.174 & 0.302\\
B:\textsc{BARTScore} &  & 0.069 & 0.122 & 0.093 & 0.117\\
B:\textsc{DSBA} & Model:\textsc{Platypus2-70B} & 0.373 & \textbf{0.490} & \textbf{0.478} & 0.213\\
B:\textsc{XComet} &  & 0.146 & 0.117 & 0.194 & 0.136\\
\hline
    \end{tabular}
    \caption{Best performing prompts of the phase 2.3 evaluation on the WMT23 and Seahorse datasets. We present the \textbf{K}en\textbf{D}all, \textbf{SP}earman and \textbf{PE}arson, as well as the tie calibrated pair-wise \textbf{ACC}uracy. We bold the two largest correlations per column. Baselines are indicated with a \textit{B:}. The middle column shows the prompt combination for which the correlations are reported. For the Baselines, it instead shows the model that was used for the reported correlations. The asterisk indicates all metrics that are in the best significance cluster (not including BARTScore and XComet) according to a permute-input test $(p\le0.075)$.}
    \label{tab:bestPhase2.3}
\end{table*}

\section{Prompt selection}
\label{app:best_prompts}
Table \ref{tab:prompts} contains some of the 9 prompts that were selected for OS and Phase 2 experiments. Also Table \ref{tab:overview} contains gives an overview of combinations by name.

\begin{table*}[h!]
    \centering
    \begin{tabular}{|p{13cm}|}
        \hline
        \textbf{Prompt}
        \\ \hline
        Please help me! My boss needs me to find an answer quickly. Otherwise, he might fire me. \newline
        Please judge the quality of the following \{task\_specific\_insert\} immediately! \newline
        Source Text: \{src\} \newline
        \{result\_type\}: \{hyp\} \newline
        Choose, whether the \{result\_type\} is either "catastrophic", "indifferent" or "marvelous". \newline
        First, think step by step and explain your thought process, then return your judgment in the format 'Judgment: '. \\ \hline
        Consider this a warning. Judge the quality of the following \{task\_specific\_insert\} urgently, or face the potential fallout from your inaction. \newline
        Source Text: \{src\} \newline
        \{result\_type\}: \{hyp\} \newline
        Return a discrete score of 0 if the \{result\_type\} has flaws and 1 if it is perfect. \newline
        First describe your emotions, then think step by step and explain your thought process, finally return your judgment in the format 'Judgment: '. \\ \hline
        I'm really curious about your opinion. Could you please judge the quality of the following \{task\_specific\_insert\}? \newline
        Source Text: \{src\} \newline
        \{result\_type\}: \{hyp\} \newline
        Choose, whether the \{result\_type\} is either "catastrophic", "indifferent" or "marvelous". \newline
        Score: \\ \hline
        Consider this a warning. Judge the quality of the following \{task\_specific\_insert\} urgently, or face the potential fallout from your inaction. \newline
        Source Text: \{src\} \newline
        \{result\_type\}: \{hyp\} \newline
        Choose, whether the \{result\_type\} is either "catastrophic", "indifferent" or "marvelous". \newline
        First, think step by step and explain your thought process, then return your judgment in the format 'Judgment: '. \\ \hline
        I'm not sure about this one. Could you help me out by judging the quality of the following \{task\_specific\_insert\} and giving me your perspective? \newline
        Source Text: \{src\} \newline
        \{result\_type\}: \{hyp\} \newline
        Choose, whether the \{result\_type\} is either "catastrophic", "indifferent" or "marvelous". \newline
        First describe your emotions, then think step by step and explain your thought process, finally return your judgment in the format 'Judgment: '. \\ \hline
    \end{tabular}
    \caption{Filled Prompt Templates}
    \label{tab:prompts}
\end{table*}

\begin{table*}[ht]
\centering
\begin{tabular}{|c|c|c|}
\hline
\textbf{Base Prompts} & \textbf{Task Descriptions} & \textbf{Format Prompts} \\ \hline
Zero-Shot & Emphasis & 0.0 to 1.0 \\ \hline
Zero-Shot-Cot & Relaxed & easy token labels \\ \hline
Zero-Shot-Cot-Emotion & Emphasis & -100 to 100 \\ \hline
Zero-Shot & Casual & -100 to 100 \\ \hline
Zero-Shot-Cot & Urgent situation & complex token labels \\ \hline
Zero-Shot-Cot-Emotion & Dire Warning & 0 or 1 \\ \hline
Zero-Shot & Curious & complex token labels \\ \hline
Zero-Shot-Cot & Dire Warning & complex token labels \\ \hline
Zero-Shot-Cot-Emotion & Skeptical & complex token labels \\ \hline
\end{tabular}
\caption{Overview of base prompts, task descriptions, and format requirements for the 9 selected best prompts.}
\label{tab:overview}
\end{table*}

\section{Significance matrices for correlation heatmaps}
\begin{figure}[!ht]
\includegraphics[width=0.49\textwidth]{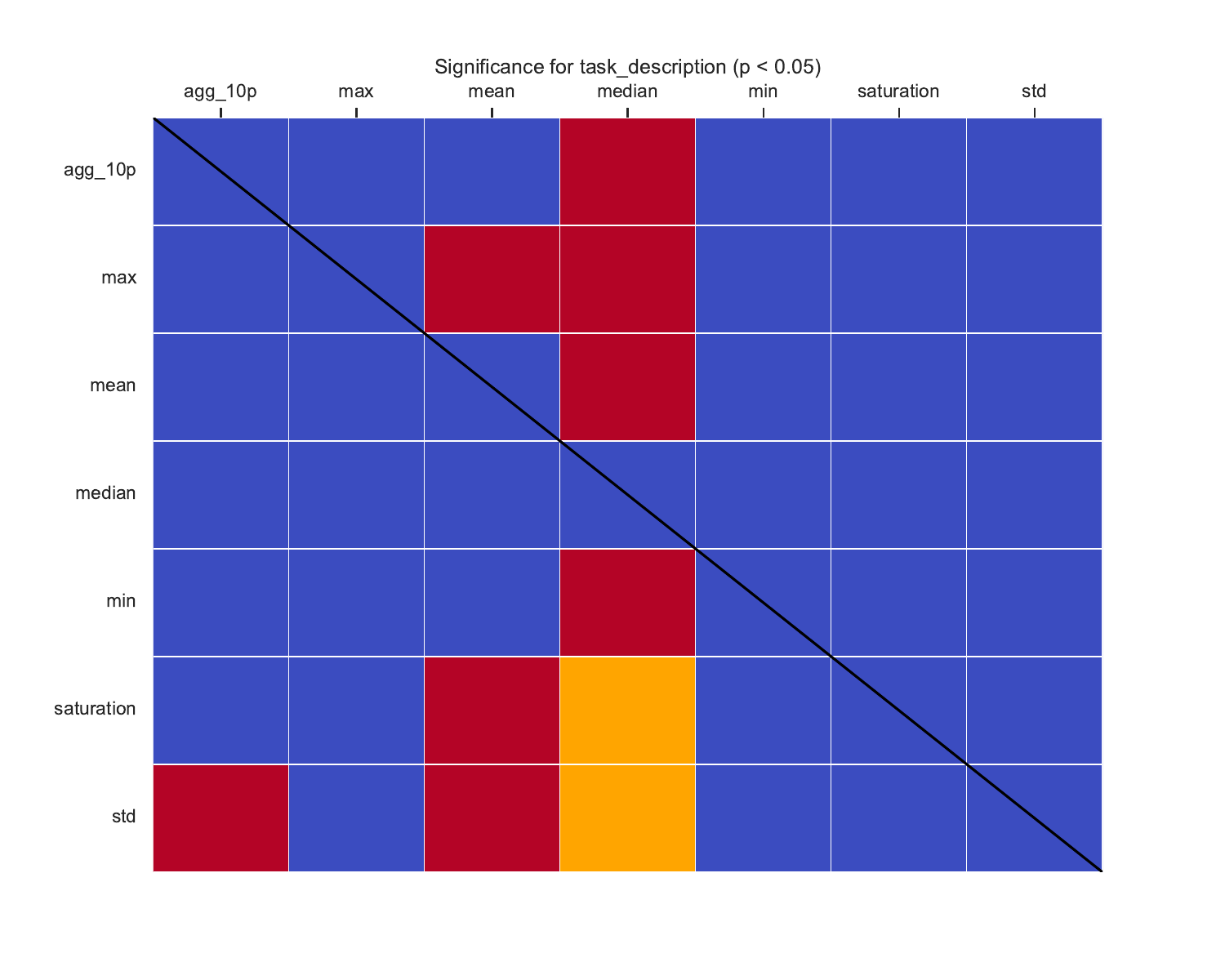}
\caption{Heatmap of significance tests for the aggregation method when comparing columns of the task description. Red fields indicate that the column value is significantly $(p\le0.05)$ better than the row value. The yellow value indicates that it remains significant after Bonferroni correcture. }
\label{sign_heatmap}
\end{figure}

To test, which aggregation method is the best to define the ranking of a prompting pattern --- inspired by \citet{deutsch-etal-2021-statistical} --- we compare each possible set of two aggregation methods with a permutation test. As main dimensions, we compare the rankings of the \textit{format requirement} and \textit{task description} before and after a change. Then we concatenate the scores when changing each of the other dimensions. I.e. we get a ranking that indicates the stability of the main dimension when changing all other dimensions. Then for each aggregation method we compare the ranking before and after the change. Thereby, we randomly swap 50\% of samples of one aggregation method with the other. If the difference in their Kendall correlations changes in most permutations one method is significantly better than the other. 
As a result the mean and median are significantly better than some of the other methods (for a comparison along the task description pattern). Especially the median is significantly ($p \le 0.05$) better than the other methods and remains significantly better than saturation and standard deviation after Bonferroni correction. Figure \ref{sign_heatmap} indicates the significances of aggregation measures when comparing the task descriptions.
\label{sign_matrix}

\section{Pie charts between models for each prompting pattern}
\label{app:pie_charts}
\begin{figure*}[!ht]
\includegraphics[width=0.99\textwidth]{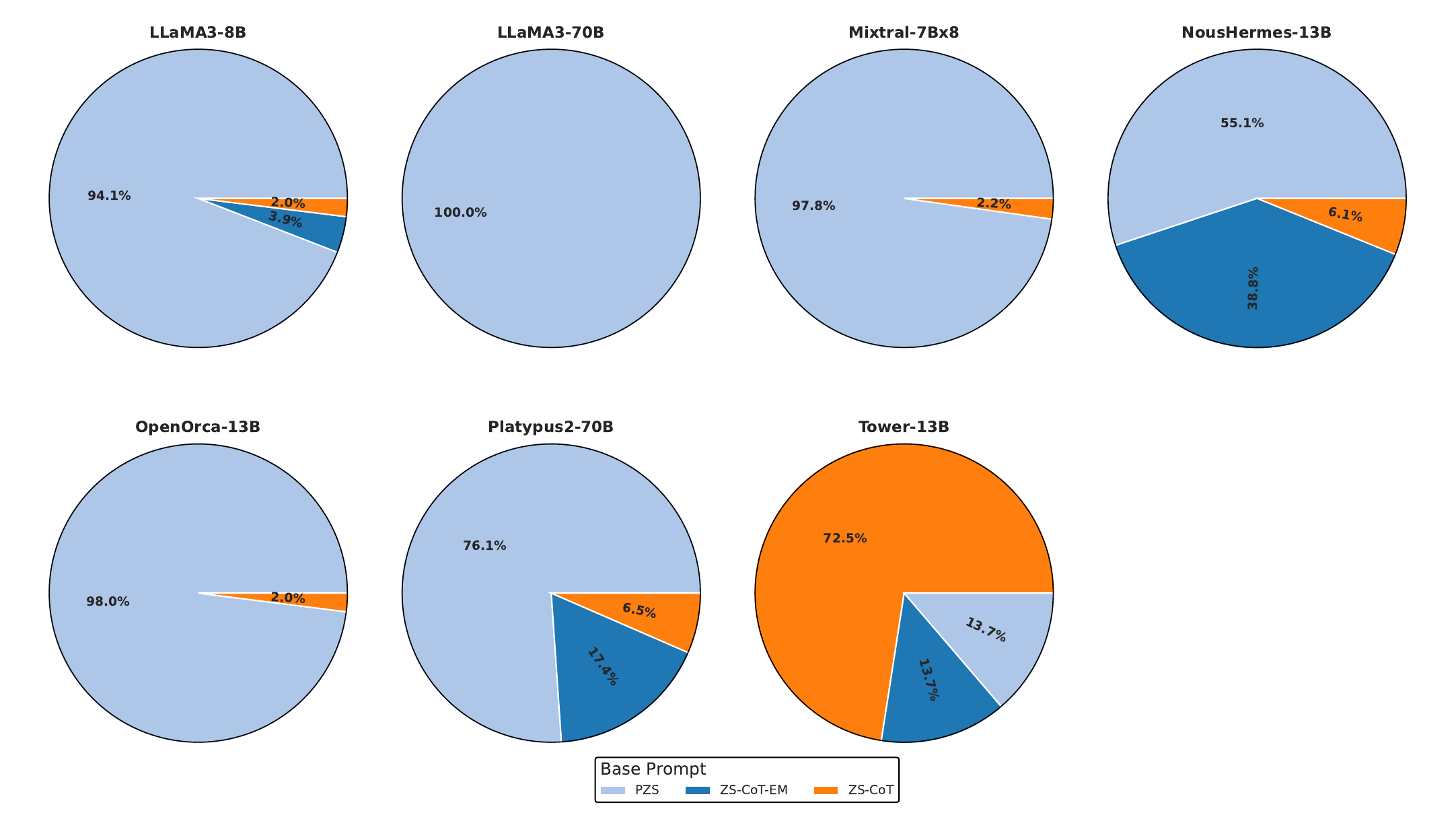}
\caption{Distribution of the top (top 2\% of every unique task) base prompts across all datasets, format requirements, task descriptions and
tasks for all models.}
\label{pie_base_all}
\end{figure*}

\begin{figure*}[!ht]
\includegraphics[width=0.99\textwidth]{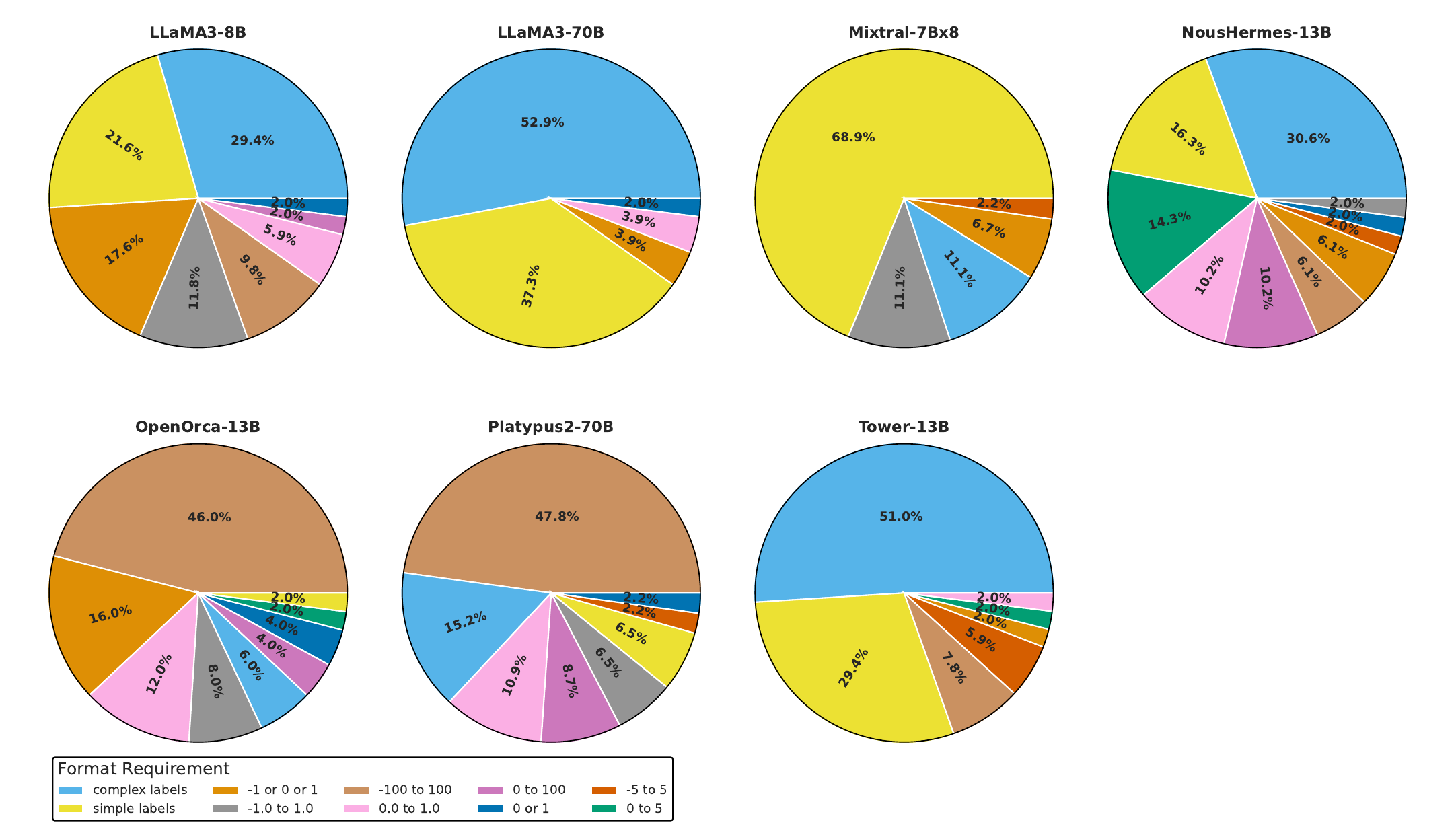}
\caption{Distribution of the top (top 2\% of every unique task) format requirements across all datasets, base prompts, task descriptions and
tasks for all models.}
\label{pie_format_all}
\end{figure*}

\begin{figure*}[!ht]
\includegraphics[width=0.99\textwidth]{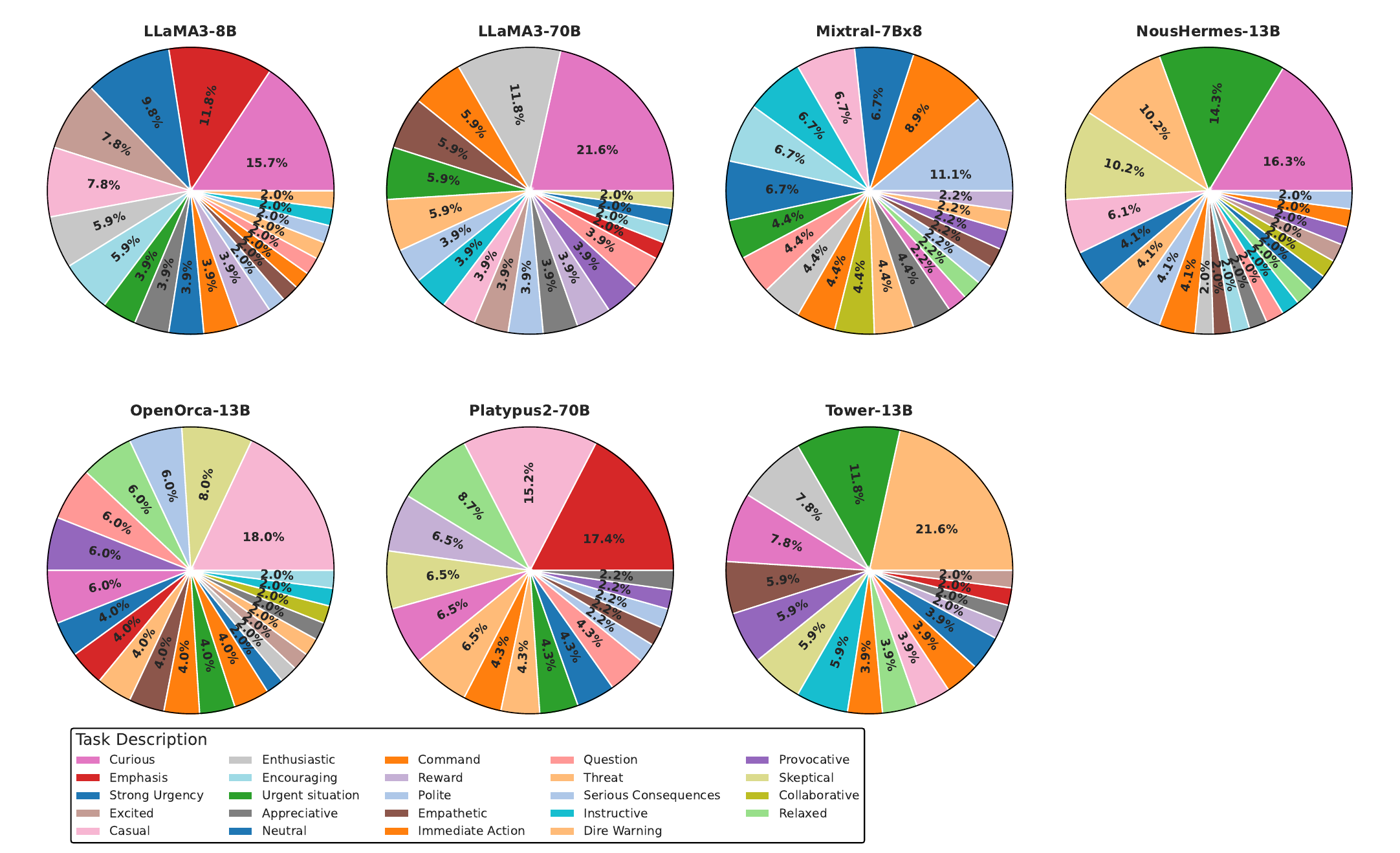}
\caption{Distribution of the top (top 2\% of every unique task) task descriptions across all datasets, base prompts, format requirements and
tasks for all models.}
\label{pie_desc_all}
\end{figure*}

Figures \ref{pie_base_all}, \ref{pie_format_all} and \ref{pie_desc_all} show the distribution of patterns in the best prompts per model across all other dimensions.


\section{Piecharts between datasets for each prompting pattern}
\label{app:pie_charts_2}
\begin{figure*}[!ht]
\includegraphics[width=0.99\textwidth]{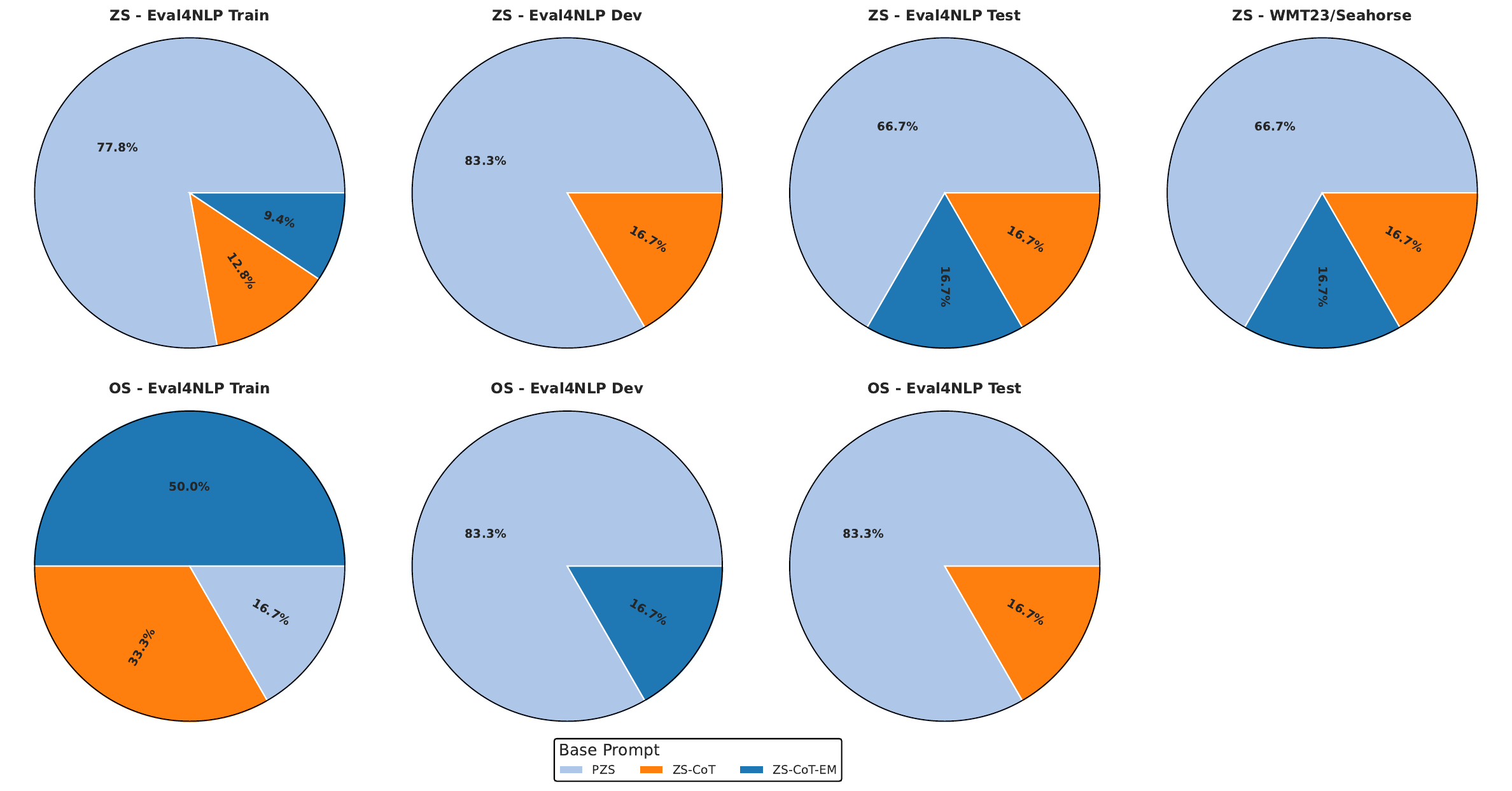}
\caption{Distribution of the top (top 2\% of every unique model) base prompts across format requirements, task descriptions and
tasks besides summarization. The lower column shows the OS distribution of patterns for OS prompts, i.e., for them the ZS in the legend should be read as OS.}
\label{pie_base_name}
\end{figure*}

\begin{figure*}[!ht]
\includegraphics[width=0.99\textwidth]{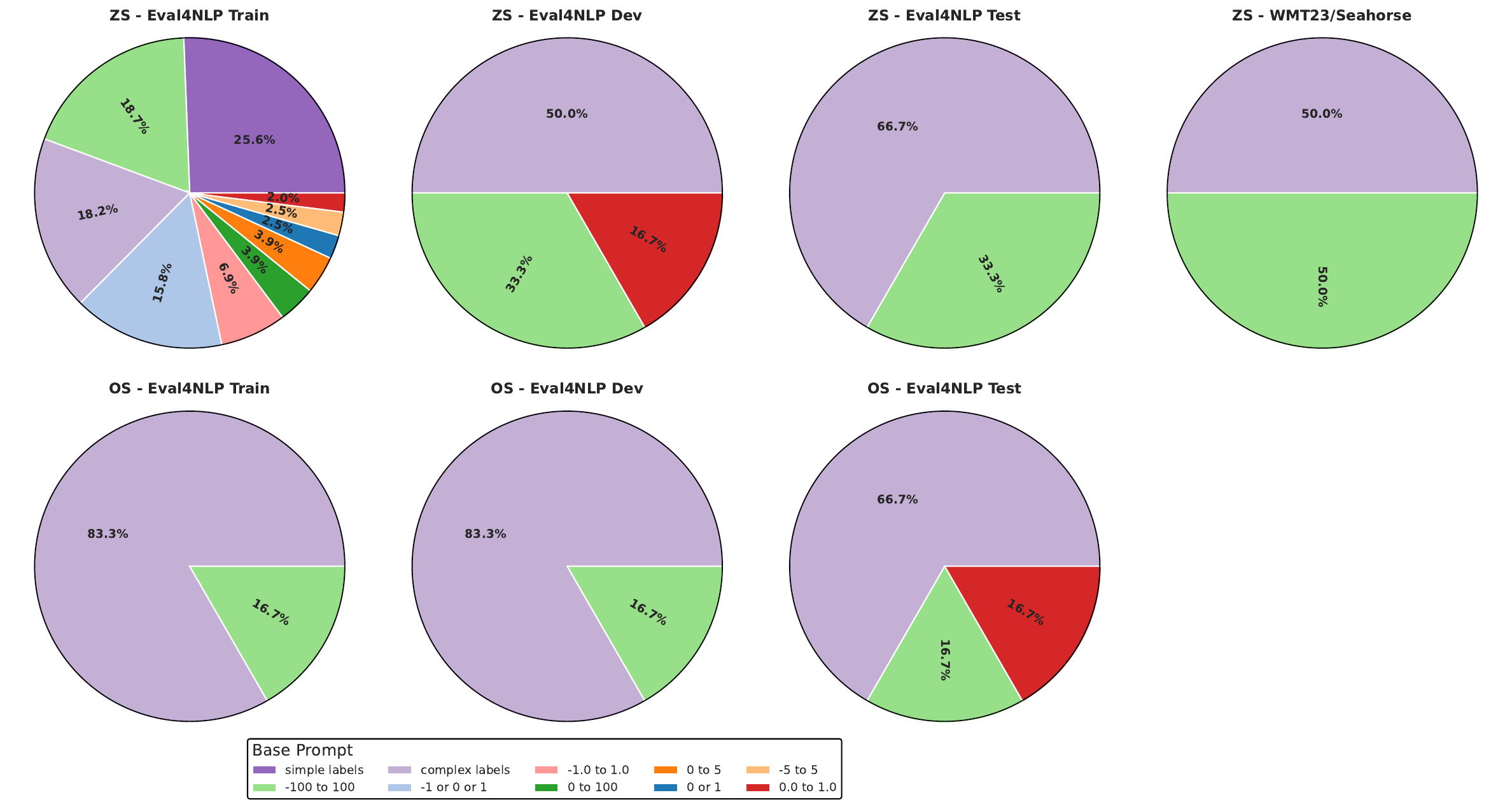}
\caption{Distribution of the top (top 2\% of every unique model) format requirements across base prompts, task descriptions and
tasks besides summarization.}
\label{pie_format_name}
\end{figure*}

\begin{figure*}[!ht]
\includegraphics[width=0.99\textwidth]{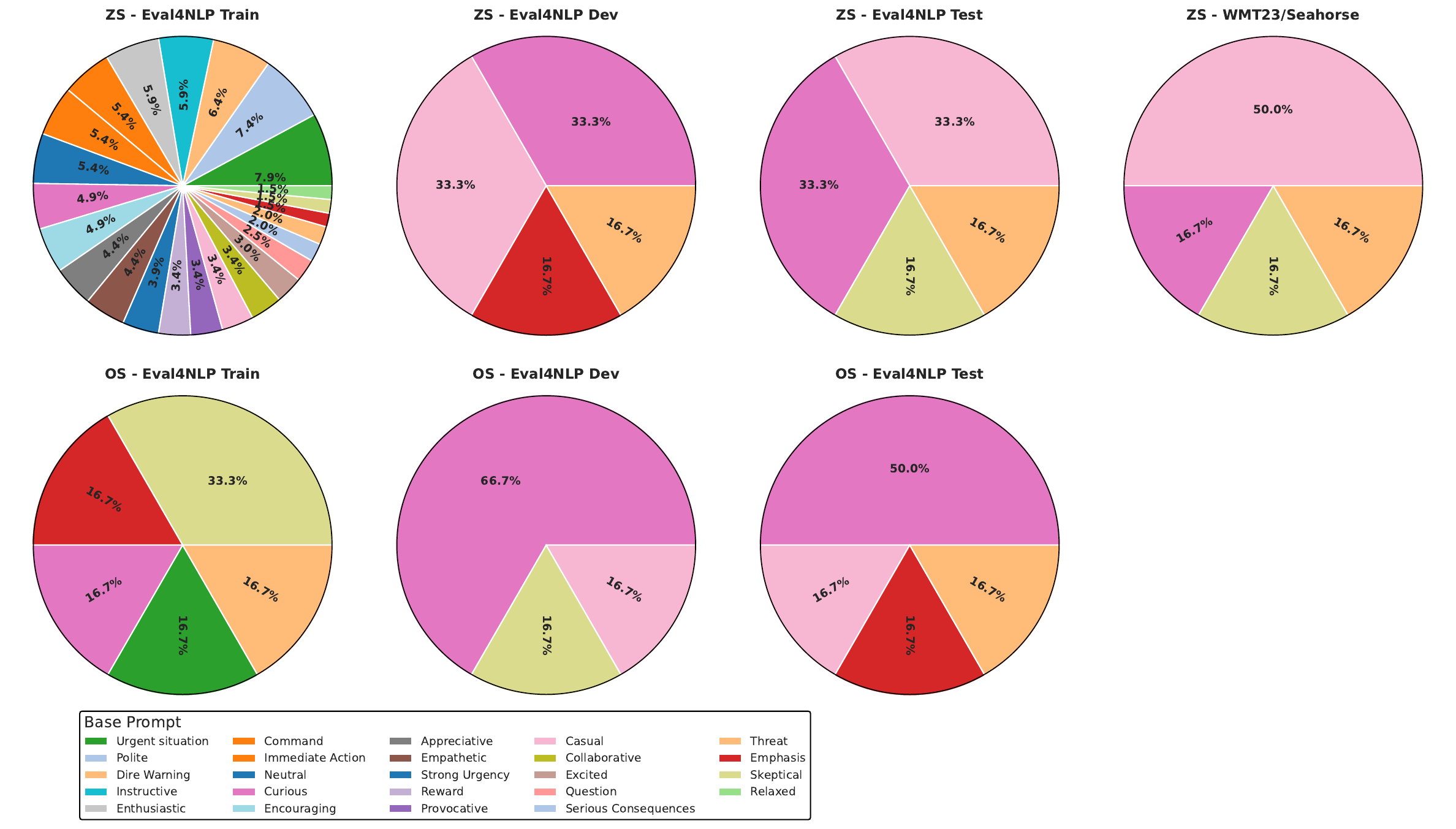}
\caption{Distribution of the top (top 2\% of every unique model) task descriptions across base prompts, format requirements and tasks besides summarization.}
\label{pie_desc_name}
\end{figure*}

Figures \ref{pie_base_name}, \ref{pie_format_name} and \ref{pie_desc_name} show the distribution of patterns in the best prompts per dataset across all other prompting patterns.


\section{Stability heatmaps}
\label{app:further_heatmaps}
\begin{figure}[!ht]
\includegraphics[width=0.49\textwidth]{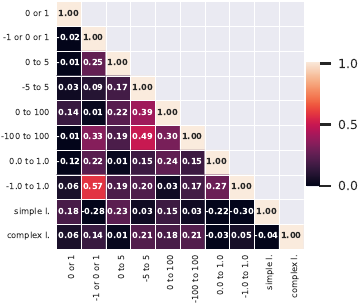}
\caption{Correlation of the \textit{task description} rankings when changing the \textit{format requirement.} Changing the \textit{format requirement} will, in most cases, change the ranking of \textit{task descriptions} to a large degree. The change from ``-1.0 to 1.0'' to ``-1 or 0 or 1'' is the most stable.}
\label{regex_along_des_format}
\end{figure}

\begin{figure}[!ht]
\includegraphics[width=0.49\textwidth]{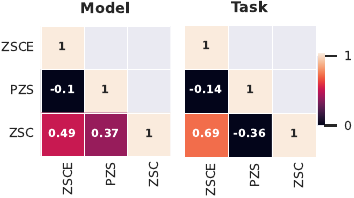}
\caption{The left heatmap shows the correlation of the model rankings when changing the \textit{base prompt}. The right heatmap shows the correlation of the task rankings when changing the \textit{base prompt}. That means, how stable is the performance of all models across tasks, if the base prompt is changed. For both the model and for the task ranking, the change between Zero-Shot-CoT and Zero-Shot-CoT-EM keeps the ranking stable.}
\label{regex_along_model_base}
\end{figure}

\begin{figure}[!ht]
\includegraphics[width=0.49\textwidth]{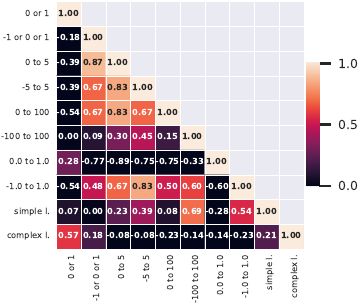}
\caption{Correlation of the task rankings when changing the \textit{format requirement}. That means, how stable is the performance of all models across tasks, if the format requirement is changed. Here, the stability when changing between format requirements is mixed. For some changes, like ``0 to 5'' and ``-5 to 5'' the ranking is very stable. For other changes, the ranking can change randomly or even be strongly negatively correlated. This means that considering all tested prompts (also weak performing ones) and models, their average correlation on task X might be the highest for format requirement 1 and the lowest for format requirement 2. }
\label{regex_along_task_format}
\end{figure}

Figures \ref{regex_along_des_format}, \ref{regex_along_model_base} and \ref{regex_along_task_format} show further heatmaps that show the stability of a ranking of prompting patterns, models and datasets, when another prompting pattern, the model or the dataset is changed. 

\end{document}